\documentclass[trans]{IEEEtran}

\usepackage{xspace,amsmath,amssymb,amsfonts,epsfig,syntonly,amsthm}
\usepackage{cite,bm,color,url,textcomp}
\usepackage{algorithmicx,algorithm,algpseudocode}
\usepackage{epstopdf}
\usepackage{empheq}
\usepackage{balance}
\usepackage{diagbox}
\usepackage{makecell, multirow, colortbl}
\usepackage{array}
\usepackage{pifont, bbding}

\newcommand*{\checkm}{\ding{52}}
\newcommand*{\cross}{\ding{56}}

\definecolor{mygray}{gray}{.9}
\definecolor{mypink}{rgb}{.99,.91,.95}
\definecolor{mycyan}{cmyk}{.2,0,0,0}
\definecolor{mygreen}{RGB}{255,215,0} 

\DeclareMathOperator*{\argmax}{arg\,max}

\long\def\symbolfootnote[#1]#2{\begingroup
\def\thefootnote{\fnsymbol{footnote}}
\footnote[#1]{#2}\endgroup}

\makeatother
\psfull

\allowdisplaybreaks[3]

\begin{document}

\title{Distributed Machine Learning for Wireless Communication Networks: Techniques, Architectures, and Applications }

\author{Shuyan Hu,~\IEEEmembership{Member, IEEE}, Xiaojing Chen,~\IEEEmembership{Member, IEEE},
Wei Ni,~\IEEEmembership{Senior Member, IEEE}, \\
Ekram Hossain,~\IEEEmembership{Fellow, IEEE}, and Xin Wang,~\IEEEmembership{Senior Member, IEEE}

\thanks{
S. Hu is with the State Key Laboratory of ASIC and System, the School of Information Science and Technology, Fudan University, Shanghai 200433, China
(e-mail: syhu14@fudan.edu.cn).

X. Chen is with the Shanghai Institute for Advanced Communication and Data Science, Shanghai University, Shanghai 200444, China
(e-mail: jodiechen@shu.edu.cn).

W. Ni is with the Commonwealth Scientific and Industrial Research Organization (CSIRO), Sydney, NSW 2122, Australia
(e-mail: wei.ni@data61.csiro.au).

E. Hossain is with the Department of Electrical and Computer Engineering, University of Manitoba, Winnipeg, MB R3T 5V6, Canada
(e-mail: ekram.hossain@umanitoba.ca).

X. Wang is with the State Key Laboratory of ASIC and System, the Shanghai Institute for Advanced Communication and Data Science, the Department of Communication Science and Engineering, Fudan University, Shanghai 200433, China
(e-mail: xwang11@fudan.edu.cn).

Shuyan Hu and Xiaojing Chen contributed equally to this work. The work of E. Hossain was supported by a Discovery Grant from the Natural Sciences and Engineering Research Council of Canada (NSERC).
The work of X. Wang was supported by the Innovation Program of Shanghai Municipal Science and Technology Commission Grant No. 20JC1416400, and the National Natural Science Foundation of China Grant No. 62071126. }
}

\maketitle

\setcounter{page}{1}




\begin{abstract}
Distributed machine learning (DML) techniques, such as federated learning, partitioned learning, and distributed reinforcement learning,
have been increasingly applied to wireless communications. This is due to improved capabilities of terminal devices,
explosively growing data volume, congestion in the radio interfaces, and increasing concern of data privacy.
The unique features of wireless systems, such as large scale, geographically dispersed deployment, user mobility, and massive  amount of data,
give rise to new challenges in the design of DML techniques. There is a clear gap in the existing literature in that the DML techniques
are yet to be systematically reviewed for their applicability to wireless systems.
This survey bridges the gap by providing a contemporary and comprehensive survey of DML techniques with a focus on wireless networks. Specifically, we review the latest applications of DML in power control, spectrum management, user association, and edge cloud computing. The optimality, scalability, convergence rate, computation cost, and communication overhead of DML are analyzed.
We also discuss the potential adversarial attacks faced by DML applications,
and describe state-of-the-art countermeasures to preserve privacy and security.
Last but not least, we point out a number of key issues yet to be addressed,
and collate potentially interesting and challenging topics for future research.
\end{abstract}

\begin{IEEEkeywords}
Distributed machine learning, wireless communication networks, convergence, computation and communication cost,
architecture and platform, data privacy and security.
\end{IEEEkeywords}

\section{Introduction}

With its capability in big data processing, adaptability to environmental dynamics, and fast speed in problem-solving,
machine learning (ML) has been increasingly applied to communication networks for improved system operation~\cite{datalife},
surveillance~\cite{traffic19, sura20}, and optimization~\cite{jing20, cuiqm20,adeep10chen,energy19xu}.
For example, deep learning (DL) has demonstrated that it can offer in-depth analysis for complex communication networks
with massive data, and provide different control schemes for different protocol layers~\cite{maoq18},
while reinforcement learning (RL) and deep reinforcement learning (DRL) can make decision and inference under unknown
and dynamically changing network conditions (e.g., channel state information)~\cite{hussain20}.
Given the dispersed or distributed nature of many wireless systems, distributed machine learning (DML) is particularly useful
under different wireless network settings for the following reasons.

\begin{itemize}
\item[$\bullet$]
The first reason is that the modern mobile devices have considerably powerful computing processor
and memory~\cite{heshuo20}. 
Their local hardware resources and data allow them to perform learning tasks in a distributed fashion.
This is critical for future wireless networks~\cite{lyu20}, in the Internet-of-Things (IoT) era, where ubiquitous connectivity will be offered with 
new applications, such as robotics, autonomous driving, and unmanned drones.
\item[$\bullet$]
The second reason for the use of DML is that the devices produce explosively large amounts of data, as a result of the diversification of services, expansion of network scale, and proliferation of wireless and mobile devices. 
The data could provide rich information and shed valuable insights, and help to significantly improve the design, deployment, operation and performance of wireless systems.

\item[$\bullet$]
Another reason for using DML is that the increasingly congested wireless channels resulting from the explosively large volume of data, and the increasing awareness of users on data privacy, discourage the user terminals from sending their data to the ML engines or servers. On the other hand, what is really critical to the ML servers is the characteristics of the data,
e.g., gradients derived from the data to update their ML models~\cite{jing20}, rather than the data itself.
\end{itemize}

For the above reasons, learning the data locally and generating the learning models and inference without the need of sending the data is of practical value. The implementation of learning at individual devices holds the key to reducing data exchange and bandwidth occupation in wireless communication networks. In particular, multiple devices can train a global ML model with their own data. The local data can participate in the training process and does not need to be directly uploaded to the server, or shared among devices.
Only local training results, e.g., the gradients of the model parameters, are sent to the server.
Therefore, DML can relieve computation and storage burden on the (central) servers,
meanwhile protecting sensitive information and preserving data privacy of the devices in wireless networks.

ML algorithms need to be integrated with DML architectures, to address large-scale ML problems. The designs of the architectures are expected to optimize the algorithm performance (e.g., high accuracy and fast convergence), and use the hardware resources efficiently. In many years of practice, researchers have developed a range of general DML architectures, such as MapReduce~\cite{White12Hadoop}, parameter server~\cite{limu13,limu14}, and graph processing architectures~\cite{Tian2018Cymbalo}. These architectures accommodate the core logic of DML algorithms and provide high-level application programming interfaces. However, {\em the applications of these architectures to wireless communications are yet to be investigated}. 

The existing surveys have been typically focused on centralized ML techniques to wireless networks,
such as Q-learning for femtocell networks and Bayesian learning for massive multiple-input multiple-output~(MIMO) systems~\cite{jiang17},
DL for different network layers~\cite{maoq18, patras19, chenmz19},
model-free strategy learning in cognitive radio environments~\cite{niyato16},
blockchain-based ML paradigms for communications~\cite{liuliu20},
and multiple ML technique for massive machine-type communications~\cite{sharma20},
edge and cloud computing~\cite{kato20}, IoT~\cite{sezer18, hussain20, lei20},
and wireless sensor networks (WSNs)~\cite{niyato14}.

The survey in~\cite{maoq18} reviews DL techniques from the perspectives of physical layer modulation and coding,
data link layer access control and resource allocation, as well as routing layer path search and traffic balancing.
In~\cite{chenmz19}, DL approaches are surveyed for emerging applications, including edge caching and computing,
multiple radio access and interference management, with an emphasis on neural networks.
ML for mobile edge computing (MEC) is reviewed in~\cite{kato20} for offloading decision, server deployment,
overhead management, and resource allocation.
DRL is summarized in~\cite{lei20} with applications to autonomous IoT networks in terms of the network layer (communication), application layer (edge/fog/cloud computing), and perception layer (physical systems).
DRL is examined in~\cite{niyato19} with a focus on applications to dynamic network access,
data rate control, wireless caching, data offloading, network security, and connectivity preservation.
In~\cite{niyato14}, the advantages and disadvantages of several ML algorithms are evaluated with guidance provided for WSN designers.
The $30$-year history of ML is revisited in~\cite{jing20} with analysis of
supervised learning, unsupervised learning, and RL for several wireless systems,
such as heterogeneous networks, cognitive radios, IoT, and machine-to-machine networks.
Moreover, none of these existing works reviewed the privacy preservation of ML-empowered communication systems.

Existing studies on DML have been heavily focused on federated learning (FL), related technologies and protocols,
and several application scenarios~\cite{yang19FL, sahu20, niknam20, lim20, rahman20}.
In particular, a tutorial on different FL structures is presented in~\cite{yang19FL}, including horizontal FL, vertical FL,
and federated transfer learning.
A tutorial on the implementation challenges of FL is provided in~\cite{sahu20}, with an emphasis on communication cost,
system heterogeneity (e.g., asynchronous communication), statistical heterogeneity (e.g., heterogeneous data), and data privacy.
FL-enabled edge computing and caching, and spectrum management methods are introduced in~\cite{niknam20},
with a discussion on system security.
FL for MEC is comprehensively discussed in~\cite{lim20} from the perspectives of cyberattack detection,
caching and computation offloading, base station association, and vehicular networks.
Extensive application scenarios of FL are illustrated in~\cite{rahman20}, such as healthcare, robotics, online retailers,
recommender system, and electric vehicles.
Yet,  the works neither provide a comprehensive summary on the DML techniques from the viewpoints of algorithm, framework,
architecture and platform, nor analyze their convergence and scalability, computation and communication efficiency,
and privacy and security under different wireless settings.

\begin{table*}[t]
\renewcommand{\arraystretch}{1.5}
\centering
\caption{Related existing surveys on ML for wireless networks and DML, and key contributions of this paper}\label{tab.review}
\begin{tabular}{ | p{1.6cm}<{\centering} | p{1.4cm}<{\centering} |  p{2cm}<{\centering} | p{1.8cm}<{\centering}  |  p{8cm}<{\centering}  |}
\hline

Related work  &Year of publication  &Main topic   &Related content in this paper &Difference and improvement of this paper \\ \hline \hline
\cite{maoq18, patras19, chenmz19}   &2018--2019   &DL for wireless communications    & Sec. II-A, ~~Sec. VIII-D 
& DML techniques and applications to wireless networks  \\ \hline
\cite{kato20}  &2020    &ML for MEC   &Sec. II-B, ~~Sec. VIII-D     &DML techniques and various application scenarios in wireless networks
\\ \hline
\cite{lei20, niyato19}   &2019--2020    &DRL for wireless networks     &Sec. II-B  
&DML techniques and various application scenarios in wireless communications   \\ \hline
\cite{yang19FL, sahu20}  &2019--2020  &Frameworks and challenges of FL  &Sec. III-A
&Other DML techniques in addition to FL, their convergence, scalability, computation and communication cost,
and countermeasures against adversarial attacks to the learning system  \\ \hline
\cite{lim20, niknam20}     &2020  &FL for MEC  &Sec. III-A, ~Sec. VIII-D   &Other DML techniques in addition to FL,
and their various application scenarios to wireless networks \\ \hline
\cite{rahman20} &2020  &Extensive uses of FL  &Sec. III-A 
&Other DML techniques in addition to FL, and their various application scenarios to wireless communications \\ \hline
\end{tabular}
\end{table*}

\textbf{Motivation of this work:}
This survey on the latest DML techniques for wireless communication networks is motivated by the following reasons.
\begin{itemize}
\item[$\bullet$] {\em A clear need of DML for wireless networks}:
Centralized ML techniques require all data to be gathered in the training system,
incurring significantly high computation and communication cost, channel congestion,
and risk of data leakage and privacy violation~\cite{lim20}.
The dispersed nature of wireless networks calls for a distributed implementation of ML and training.
\item[$\bullet$] {\em Practical feasibility of DML}: This is a result of the increased capability of wireless devices.
The increasing computation power and resources of mobile devices make DML possible in practice~\cite{gu19}.
\item[$\bullet$] {\em An existing gap in the literature}: DML has not been systematically studied, reviewed,
or compared in the context of wireless systems.
Existing surveys reviewed centralized ML techniques to wireless networks~\cite{patras19, chenmz19,
niyato16, liuliu20, sharma20, kato20, sezer18, lei20},
or examined FL as the only distributed implementation of ML~\cite{yang19FL, sahu20, niknam20}.
This survey aims to bridge the gap with a focus on DML for wireless networks. 
\end{itemize}

\textbf{Contributions of this work:}
This paper presents a contemporary and comprehensive survey of DML techniques designed for wireless communication networks
to bridge the gap of the existing literature,
including the techniques of DML and their applicability to wireless communication networks, their architectures and platforms,
computation and communication efficiency, and approaches for data privacy and system security.
We review popular applications of DML in wireless communications.
In particular, our interests are in power control, spectrum management, user association, and edge cloud computing,
based on a large number of recently published results.
We analyze the optimality, scalability, convergence rate, computation and communication cost of recent algorithms and frameworks.
Given the paramount importance of security, privacy and reliability in DML,
we also discuss the potential adversarial attacks faced by servers and agents,
and introduce multiple countermeasures to preserve data privacy and enhance system security.

This survey also points out a number of key issues yet to be addressed,
and collates potentially interesting and challenging topics for future research.
Our study reveals that DML has been extensively considered to manage the resources and control the transmit powers
of wireless communications.
Tremendous amount of effort has been to date devoted to decentralizing originally centralized ML models.
For example, the majority of published results are on FL.
In contrast, multi-agent RL rooted in a game-theoretic understanding of distributed multi-agent operations
and benefiting from the adaptivity of (deep) RL, is a fast-growing area with many distributed applications to wireless resource allocation,
protocol coexistence, and cooperative communications.
The topics of related works and the key contributions of this work are summarized in Table~\ref{tab.review}.

\begin{figure}[t]
\centering
\includegraphics[width=0.49\textwidth]{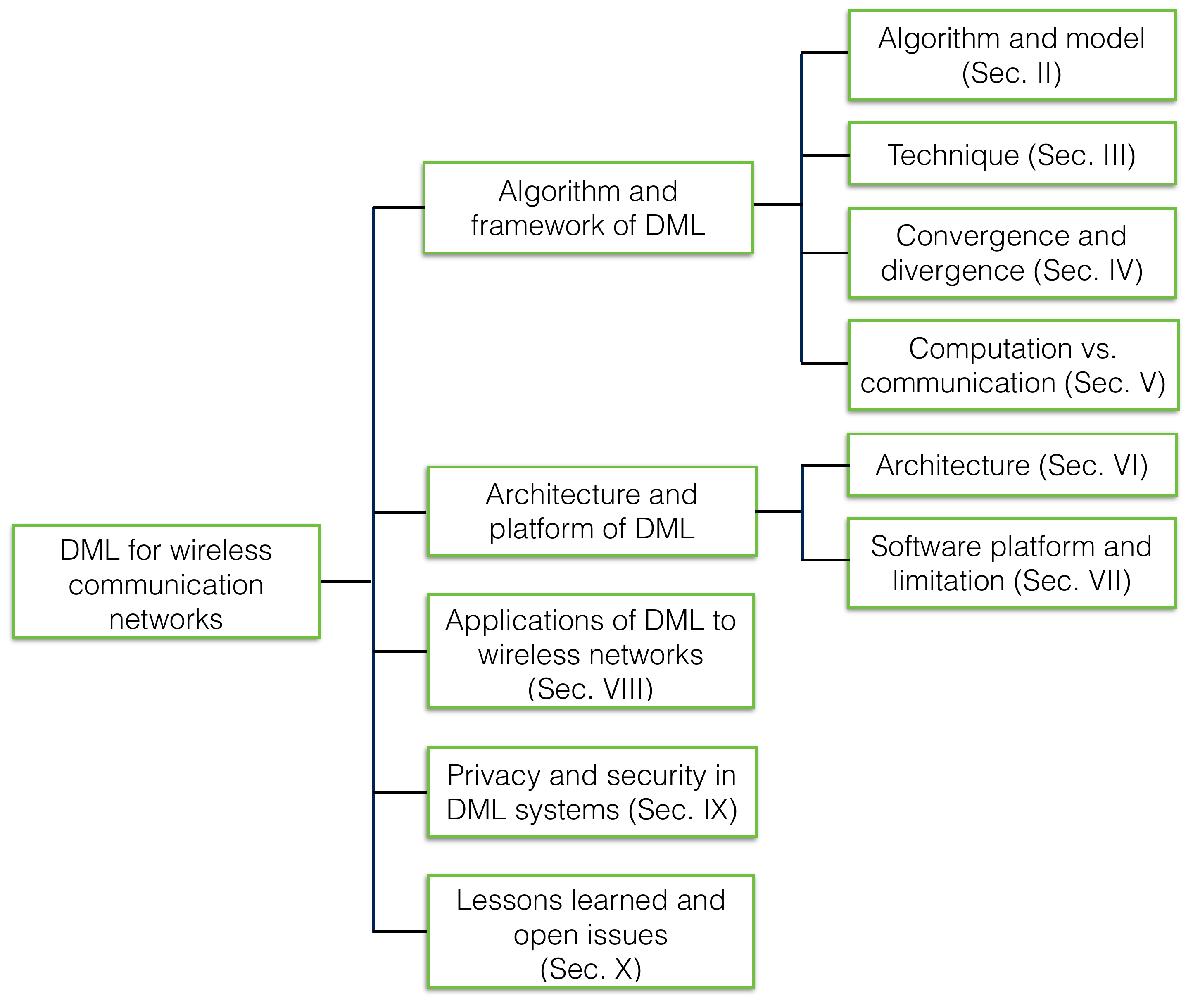}
\caption{The anatomical organization of this survey.}
\label{organize}
\end{figure}

The rest of this paper is organized as follows.
Section II provides preliminaries on DML algorithms and models.
Section III introduces DML frameworks and techniques, with their applications to wireless networks.
Section IV discusses the optimality, scalability, and convergence rate of popular algorithms, with an analysis on the gap between
input and output data.
Section V considers the computation and communication cost of recent DML algorithms and frameworks.
Sections VI and VII review the architecture and platform for DML, respectively.
Section VIII reviews the applications of DML to wireless networks, including power control, spectrum management,
user association, and edge cloud computing.
Section IX describes the state-of-the-art techniques of preserving privacy and enhancing security in DML systems
from the perspectives of both the server and the agent.
Section X summarizes the lessons learned from existing works, discusses the key issues to be addressed,
and suggests interesting and challenging topics for future research.
Fig. \ref{organize} depicts the organization of this survey.

\section{Algorithms and Models}

\begin{figure*}
\centering
\includegraphics[width=0.8\textwidth]{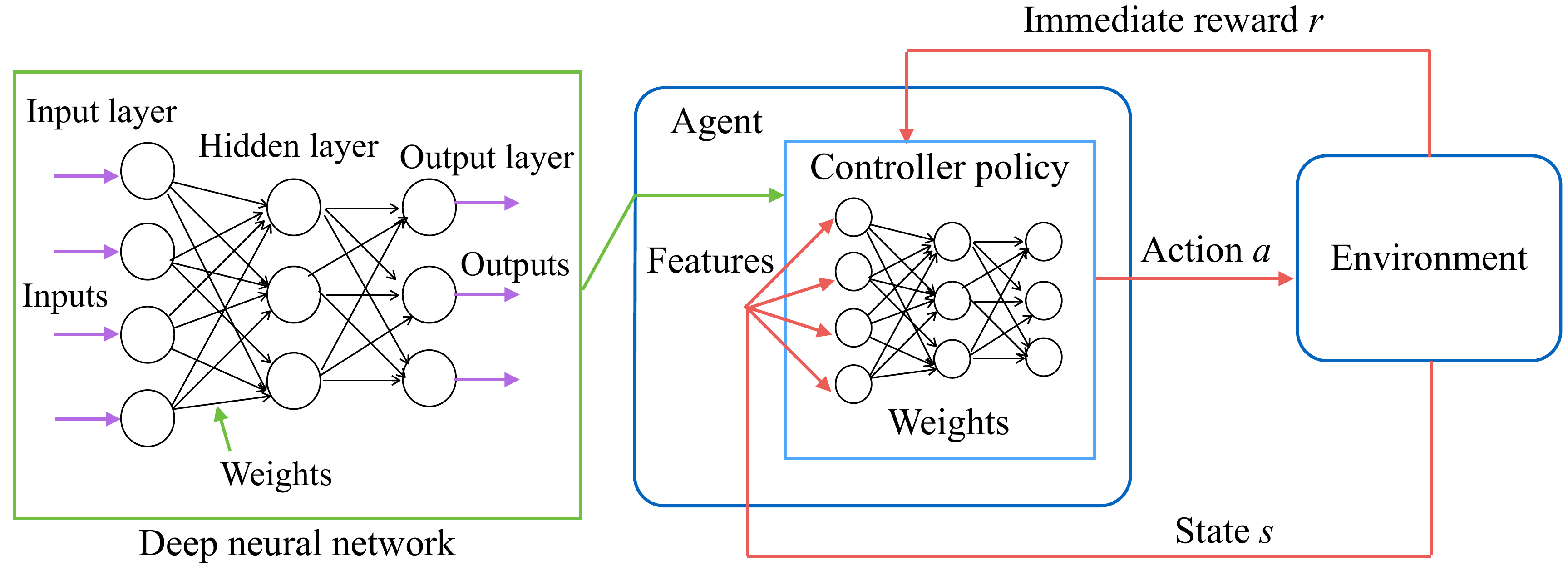}
\caption{An illustration of a deep neural network and the deep reinforcement learning process,
where an agent (i.e., NN) continuously communicates with an environment and acquires reward information as feedback.
The agent chooses an action at every step, which will alter the environment state.
The learning mission of the NN is to optimally configure its parameters,
so that it can choose actions that most likely result in the best return in the future.}
\label{DRL}
\end{figure*}

\subsection{Deep Learning (DL)}
DL consists of a collection of algorithms and methodologies that are used to model and extract
important features of data~\cite{deeplearn16}.
DL is typically applied to avoid manual description of data (for example, handwritten characteristics)
by learning automatically from the data.
Unlike its traditional ML counterparts that largely depend on attributes determined by domain experts,
DL approaches obtain information from raw data layer-by-layer via nonlinear processing units,
and make predictions based on specific objectives.
The most popular DL models are neural networks (NNs), and only those NNs with enough hidden layers
(usually, at least two layers) are considered ``deep''.
Other multi-layer architectures, for instance, deep Gaussian process~\cite{teng18sensor}, neural process~\cite{boss17},
and deep random forest~\cite{samya16}, are regarded as DL structures.
An advantage of DL compared to conventional ML is spontaneous feature extraction,
through which hand-crafted efforts can be avoided~\cite{xu19spl}.

Deep neural networks (DNNs) approximate complex functions by using simple non-linear
operations on input data.
The operations are carried out by ``neurons" which perform a weighted summing of data followed  by a non-linear transformation.
The neurons are organized into different layers ``layers'' (Fig.~\ref{DRL}).
The structure of a DNN is similar to a human brain, where particular units are activated and they influence the output of the DNN model.
The loss function of DNNs is usually differentiable.
By using stochastic gradient descent (SGD) approaches and back-propagation mechanism, which obeys the basic chain principle of differentiation, the model parameters are obtained when the minimum value of a loss function is reached~\cite{nature86}.


\begin{algorithm}[t]
\caption{The Q-learning algorithm \cite{niyato19}.}
\label{alg.qq}
\begin{algorithmic}[1]
\State {\bf Initialization:} For every state-action pair $(s,a)$, set the Q-value $Q(s,a)$ to zero.
Obtain the present environment state $s$, set a value for the training rate $\alpha$ and the discount factor $\gamma$.
\For {$t$ = 1, $\cdots$, $T$}
\State Obtain immediate reward $r$ and a new state $s'$ based on the current $(s,a)$.
\State Select an action $a'$ according to $s'$ and update the Q-value $Q(s,a)$ by:
\begin{equation}
\begin{aligned}
Q_{t+1}(s,a) \leftarrow & Q_{t}(s,a)+\alpha_t \big[r_t(s,a)+ \notag \\
&\gamma \max_{a'} Q_{t}(s' ,a')-Q_{t}(s,a) \big].
\end{aligned}
\end{equation}
\State Update $s \leftarrow s'$.
\EndFor
\State {\bf Output:} $\pi^* (s)=\argmax_a Q^*(s,a)$.
\end{algorithmic}
\end{algorithm}

\subsection{Reinforcement Learning (RL)}
RL learns to control a system to maximize a long-term objective~\cite{csaba10}.
A controller receives the state of the controlled system and a reward associated with the latest state transition.
Accordingly, the controller decides an action and returns the decision to the system.
In response, the system transits to a new state.
This cycle repeats until the controller learns a way of controlling the system to maximize the total reward.

In an ML system, an ``agent'' refers to the learning participant whose behavior in an environment can be improved through training.
Categorizing by the number of agents in a system, there are single-agent RL and multi-agent RL.
The model of the single-agent RL is captured by a Markov decision process.
The mission of the agent is to achieve the best long-term performance (return),
while only acquiring update on its immediate one-step performance (reward).
The single-agent case is generalized to the multi-agent RL by a stochastic game,
where the state transitions stem from the collaborative action of all the agents.
The rewards of the agents rely on the collaborative action, and their returns rely on mutual policy.
If the reward functions and the returns are the same for all the agents,
the stochastic game is \emph{fully cooperative} in that the agents share a unified intention to maximize the mutual return.
Otherwise, the game is not fully cooperative and the agents may even have opposing goals.

The most popular type of RL is Q-learning, where an agent intends to obtain the largest value of a long-term target by communicating and cooperating with its environment according to a trial-and-error strategy in the absence of any previously-stored dataset~\cite{watkins92}.
Q-learning can acquire an adequate strategy by refreshing an execution value, i.e., the $Q$ value, without a model of the environment.
The Q value, denoted by $Q(s,a)$, is defined as the expected accumulative premium
when an execution $a$ is carried out under the environment state $s$ and the operation strategy $\pi$~\cite{watkins92, yu19}.
Once the $Q$ values are acquired after a sufficiently long period, the agent can perform the best action (i.e., the action with the best $Q$ value) at the present state~\cite{sun19}.
The Q-learning algorithm is summarized in \textbf{Algorithm~\ref{alg.qq}}.

To make the RL algorithms more capable for generalization, DRL was proposed by DeepMind~\cite{deepmind},
which consists of a group of approaches that calculate value functions (deep Q learning)
or policy functions (policy gradient scheme) via DNNs.
An agent (i.e., NN) continuously communicates with an environment and acquires reward information as feedback.
The agent chooses an action at every step, which will alter the environment state.
The learning mission of the NN is to optimally configure its parameters,
so that it can choose actions that most likely result in the best return in the future.
DRL is usually based on the SGD methods~\cite{sgdli18},
and can be readily applied to problems that have a large number of potential states (i.e., high-dimensional environments).
The framework of DRL is illustrated in Fig. \ref{DRL}.

\subsection{Stochastic Gradient Descent (SGD)}
SGD and its variants are popular techniques to train DML models.
SGD enables the output data of an ML model to approach the true distribution of the input data
by minimizing their divergence (or gap) measured by a \emph{loss function}.
In essence, SGD selects a sample $i$ for variable $x$ from the set $\{1,2, \ldots, n\}$ uniformly and randomly.
The SGD updates the variable $x$ using the gradient of the loss function by sample $i$, i.e., $\nabla f_i(x)$, which is
a stochastic estimation of the loss function by all samples, i.e., $\nabla F(x)$~\cite{sgdkone16, sgm17}.

To enhance the global optimality and convergence, multiple variants of the SGD have been proposed.
For instance, momentum-based SGD adds a weighted sum of the previous gradients as momentum to the current gradient,
which reduces fluctuations by limiting the influence of the current gradient.
Momentum-based SGD can realize the global optimality and enhance the convergence
over the first-order gradient-based methods~\cite{liu19mgd}.
Parallel mini-batch SGD (PMSGD) splits the global dataset into mini-batches, and updates them in parallel per iteration simultaneously,
which can speed up the convergence of traditional SGD~\cite{dikel12}.
Decentralized parallel SGD improves the PMSGD by requiring only a local collection of gradients between neighbors,
instead of a global aggregation~\cite{assran18}.

Recently, the Alternating Direction Method of Multipliers (ADMM) has been developed  as a plausible alternative 
for large-scale ML problems, in addition to SGD used in DL or partitioned learning.
ADMM decomposes a problem into several parallel subproblems and orchestrates overall scheduling across the subproblems to solve the original problem~\cite{iut16}.
It can guarantee a linear convergence rate under certain conditions.
Detailed discussions will be provided in Section \ref{sec.converge}.

\section{Techniques for Distributed Machine Learning}

\begin{figure}[t]
\centering
\includegraphics[width=0.43\textwidth]{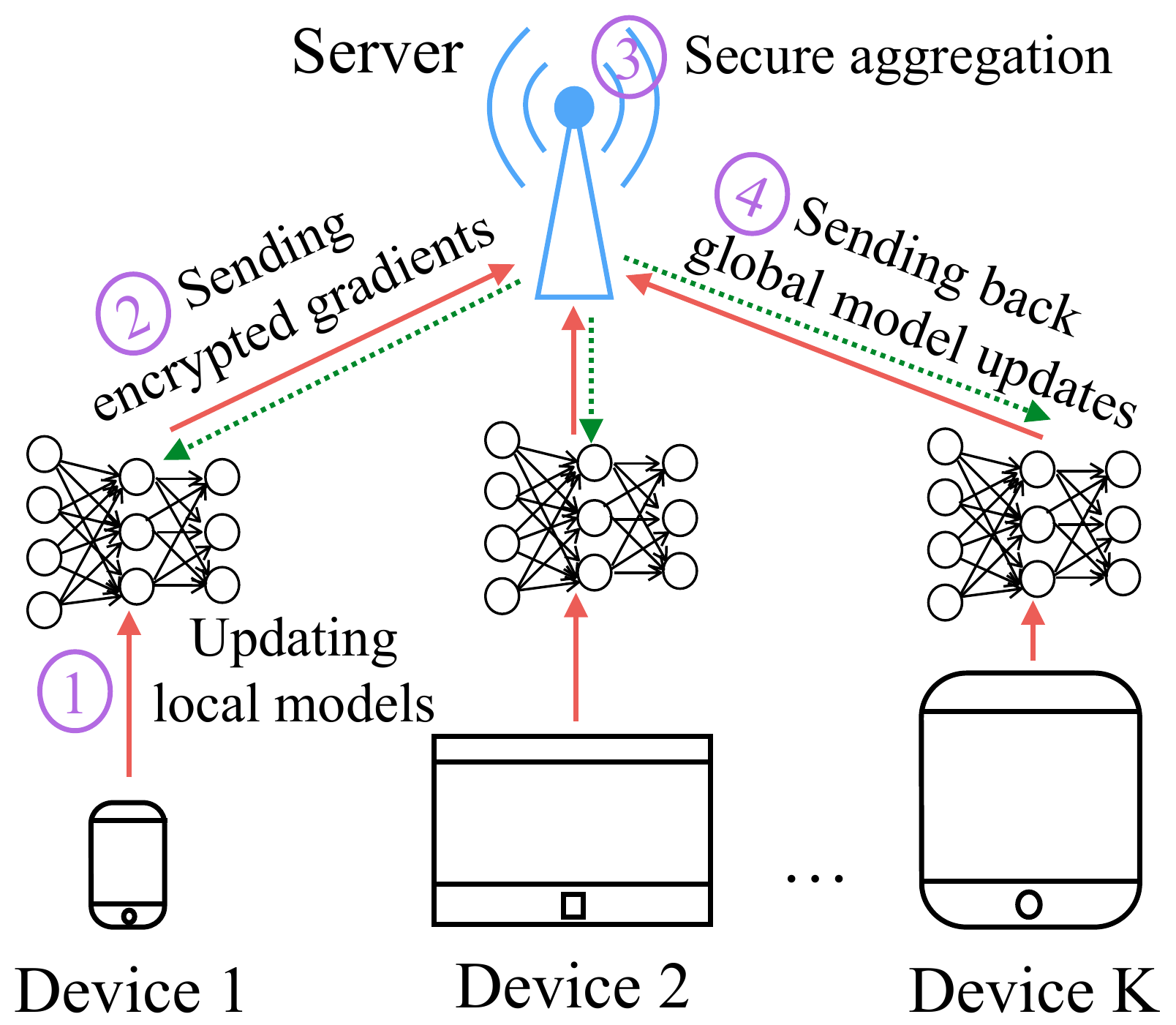}
\caption{An illustration of the horizontal FL framework.
Two distant BSs with similar communication services may have different and non-overlapping groups of users
from their respective coverage areas.
The feature spaces  are the same across the coverage areas.
A mobile terminal can update the model parameters and upload them to the server.
In this way, a global model can be established with the assistance of other mobile terminals.}
\label{horizonFL}
\end{figure}

\begin{figure*}
\centering
\includegraphics[width=0.7\textwidth]{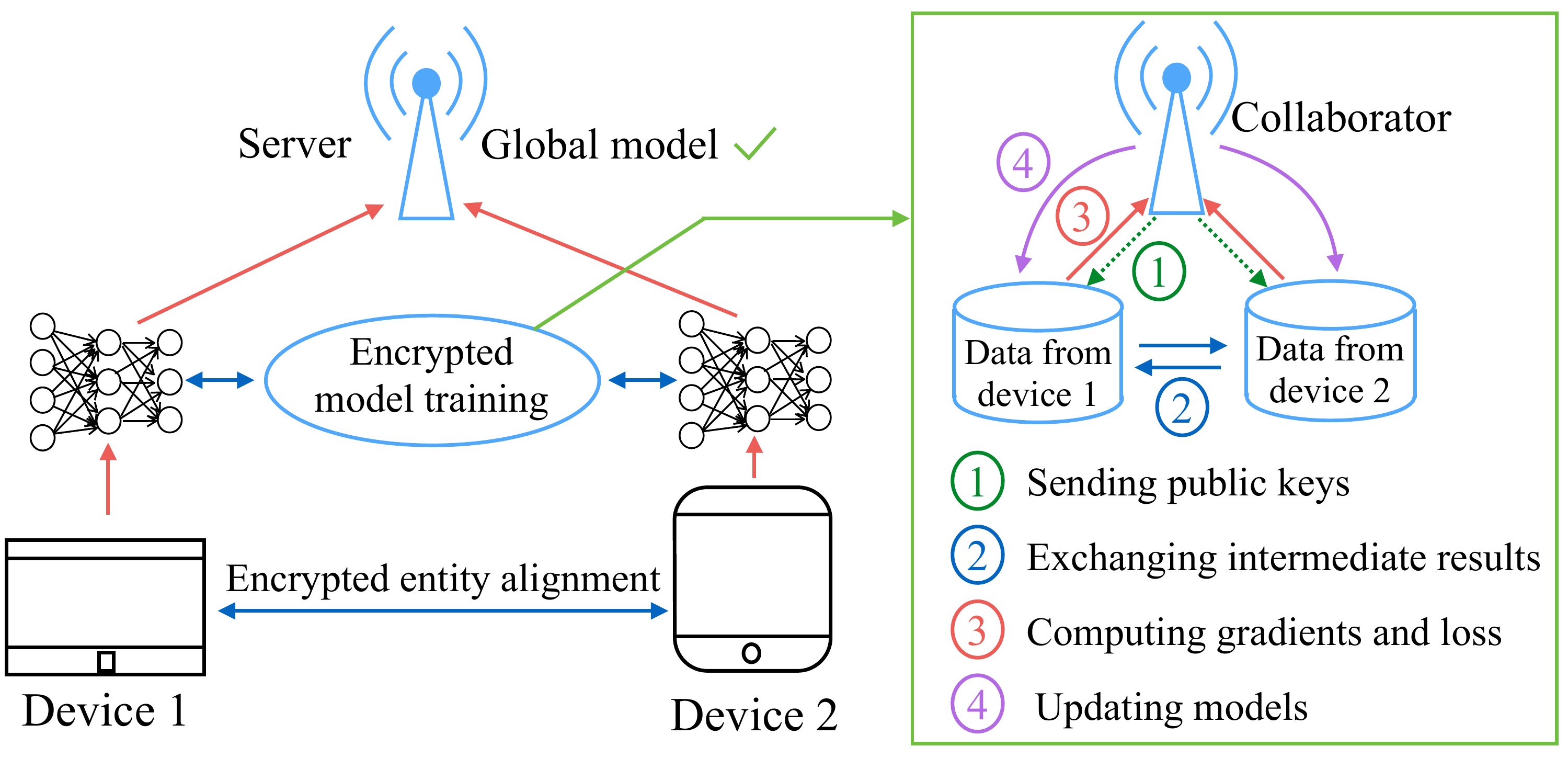}
\caption{An illustration of the vertical FL framework.
A BS and a cloud server located in the same region with the same group of users record different user behaviors,
such as communication among users through the BS, and data uploading and storage by the cloud server.
The feature spaces, collecting user behaviors, are different between the BS and the server.}
\label{verticFL}
\end{figure*}

\begin{figure*}[t]
\centering
\includegraphics[width=0.8\textwidth]{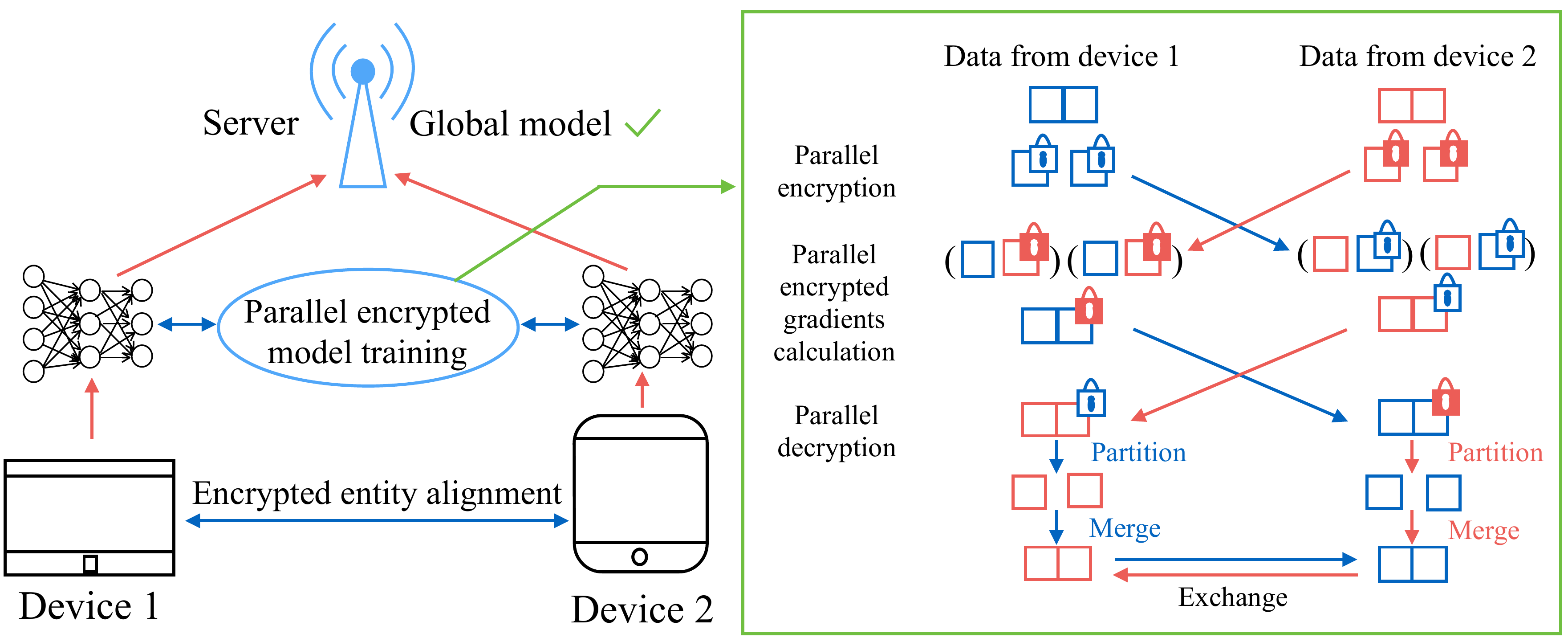}
\caption{An illustration of the (encrypted) federated transfer learning framework,
which is utilized when data sets differ in both the samples and the feature space, e.g.,
mobile user subscriptions to two different wireless operators.
Due to geographical restrictions and different service types,
a small portion of user groups and feature spaces overlap between the two entities.
By applying the FTL technique, a mutual depiction between the two characteristic spaces is obtained with the scarce public sample sets
and then used to predict samples with single-sided characteristics.}
\label{FTL}
\end{figure*}

\subsection{Federated Learning (FL)}
FL is a concept created recently by Google~\cite{kone15, jakub16},
which builds (in most cases, supervised) ML models using datasets locally available to different devices and eliminates the need for data exchange between the devices or beyond.
FL is usually trained based on SGD~\cite{sgdjustin17, sgdli18}
or a Federated Averaging (FedAvg) algorithm~\cite{mahan17}, which is summarized in \textbf{Algorithm \ref{alg.fedavg}}.
For FedAvg, every client runs a cycle of the gradient descent based on its own data to update its own ML model parameters.
Next, the server averages the model parameters of all clients to update $x$.
More computations can be added to every client by iteratively updating the local model parameters multiple times before the averaging step.
Based on the distribution characteristics of data, typical FL algorithms are divided into the following three categories~\cite{yang19FL}.

\begin{algorithm}[t]
\caption{Federated Averaging (FedAvg)~\cite{mahan17}.}
\label{alg.fedavg}
\begin{algorithmic}[1]
\State {\bf Define:} Local mini-batch size $B$, number of local epochs $E$, number of participants $N$,
learning rate $\alpha$, global model $\mathbf w_G$, and loss function $L(\mathbf w)$.
\State {\bf LocalTrain}$(i, \mathbf w)$ (at participant $i$):
\State Split local dataset $D_i$ to mini-batches ${\cal B}_i$, each in the size of $B$.
\For {Local epoch $j$ = 1, $\cdots$, $E$}
\For {$b \in {\cal B}_i$}
\State $\mathbf w \leftarrow \mathbf w - \alpha \nabla L(\mathbf w;b)$ ~($\nabla L$ is the gradient of $L$ on $b$).
\EndFor
\EndFor
\State ~
\State {\bf Training at the server:}
\State Initialize $\mathbf w_G$.
\For {Iteration $t$ = 1, $\cdots$, $T$}
\State Randomly choose $m$ participants from the total participants.
\For {Each chosen participant $i$ \textbf{parallelly}}
\State $\mathbf w_i^{t+1} \leftarrow$ \textbf{LocalTrain$(i, \mathbf w_G^t)$}.
\EndFor 
\State {\bf Output:} $\mathbf w_G^t = \frac{1}{\sum_{i=1}^N D_i} \sum_{i=1}^N D_i \mathbf w_i^t$~(Averaging aggregation).
\EndFor 
\end{algorithmic}
\end{algorithm}

\subsubsection{Horizontal FL (or Sample-Based FL)} This category is applied, when datasets share the same feature space but differ in the samples~\cite{li20horizon}.
An example application scenario of the horizontal FL is in wide areas with  uniformly deployed BSs and uniformly distributed users, e.g., low-to-medium density, suburban areas. Each user can benefit from the data collectively captured by all users, to improve its uplink power control, modulation-and-coding scheme (MCS), quality-of-service (QoS) parameters, etc. A global ML model can be trained collectively by the users. The feature spaces are the same across the users.
In a typical framework of horizontal FL, every individual mobile terminal can update the parameters of the ML model and upload the parameters to the server (or cloud).
In this way, the global ML model can be established with the assistance of individual data owners,
e.g., other mobile terminals.
The horizontal FL framework is illustrated in Fig.~\ref{horizonFL}.

\subsubsection{Vertical FL (or Feature-Based FL)} This category is employed when datasets of individual data owners are sampled from the same data distribution but can have distinct features~\cite{liu20vertical, feng20vertical, yang19vertical, yangsw19}.
An example application scenario of vertical FL is in cellular systems, where BSs and the  gateways in the cellular core network  train collaboratively a global ML model to make decisions on user association, handover, resource allocation and downlink power control. The feature spaces of the BSs include the received signal strengths, channel conditions and QoS of the users. The feature spaces of the gateways capture inter-BS features, which account for the user association, handover and interference mitigation aspects of the global ML model.
Vertical FL is applied to aggregate the distinct features and calculate the loss function of the training function and the gradients,
protecting privacy at the same time.
This allows a model to be constructed with data drawn from both sides cooperatively.
In this FL paradigm, the specific identity and state of each engaging party cannot be differentiated,
and the FL system assists every party to achieve a ``commonwealth'' policy.
The framework of vertical FL is illustrated in Fig.~\ref{verticFL}.

\subsubsection{Federated Transfer Learning (FTL)} This category is utilized when data sets differ in both the samples and the feature space~\cite{liu18ftl, sharma19}.
Take mobile user subscriptions to two different wireless operators (even in two different countries) as an example.
Due to geographical restrictions and different service types, a small portion of user groups and feature spaces, such as power control, MCS, and QoS, overlap between the two entities in this example.
By applying the FTL technique, a mutual depiction between the two characteristic spaces is obtained with the scarce public sample sets
and then used to predict samples with single-sided characteristics.
Moreover, the model trained for one of the operators can be readily transferred and applied in the network of the other operator,
when a user roams from the former operator to the latter.
The FTL technique tackles problems that exceed the scope of the horizontal and vertical FL techniques described earlier.
The framework of FTL is illustrated in Fig.~\ref{FTL}.

For the purpose of privacy preservation, a distributed (or federated) DL framework is proposed in~\cite{shokri15},
which allows several participating agents to cooperatively acquire a precise NN model for a predefined target with no need to share their input datasets.
Based on the asynchronous SGD algorithm, a paralleled and distributed learning process among the participants is enabled
through a parameter server.
In particular, each participant learns by itself using its own dataset.
The participant chooses to expose only its core parameters in part (i.e., gradients) to refresh the entire model.
While benefiting from the models of the others, each participant can protect its sensitive data and privacy,
and achieve the convergence and learning accuracy of the global model.

Horizontal FL has been widely applied to wireless networks for power control \cite{tung20}, spectrum management \cite{kaibin19},
edge cloud computing \cite{xujie20},
since it allows for distributed training and update of a global model at each device without data exchange
between the devices.
Despite the fact that vertical FL and FTL have yet to be implemented in wireless applications, their capability of model training with distinct data features
indicates that they could have great potentials in large-scale geographically distributed systems
and achieve desirable global learning performance.
Detailed discussions of these applications will be provided in Section \ref{sec.use}.

\subsection{Partitioned Learning}
\begin{figure}[t]
\includegraphics[width=0.47\textwidth]{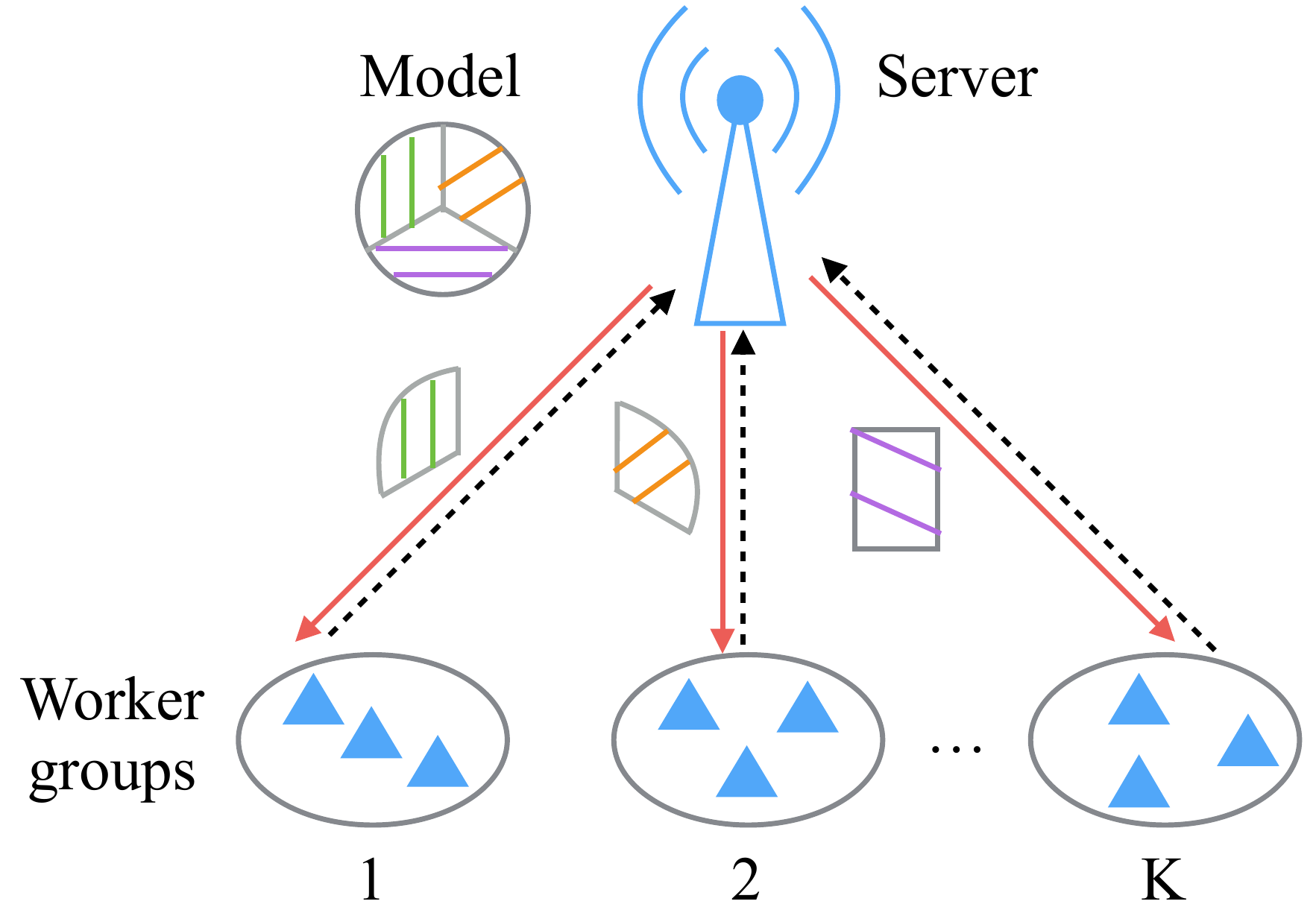}
\caption{An illustration of the partitioned learning framework, where a supervised learning model is broken down into multiple blocks.
Each block contains different parameters and is downloaded to different devices for decentralized computing.
The updated mathematical outcomes are returned to a server to refresh the entire model.}
\label{partition}
\end{figure}

Under the setting of partitioned learning, a supervised learning model is broken down into multiple blocks.
Each block contains different parameters and is downloaded to different devices for decentralized computing.
The updated mathematical outcomes are returned to a server to refresh the entire model~\cite{wen20}.
A well-known framework of partitioned decentralized learning is the \emph{Parameter Server} framework~\cite{limu13, limu14}
described in Section \ref{sec.ps}.
Applying the \emph{block coordinate descent} (BCD) method~\cite{wright15}, the Parameter Server framework
decomposes a large-scale model-optimization problem into a separable target function,
e.g., linear regression~\cite{geng19tsp} and support vector machine (SVM)~\cite{song19svm},
and solves the problem iteratively in a decentralized manner.
The framework of partitioned learning is illustrated in Fig. \ref{partition}.
The key difference of the partitioned learning to FL is that FL does not perform model partitioning
and demands edge devices to jointly renew the entire model.
The properties of different DML techniques are summarized in Table \ref{tab.feat}.

\begin{table*}[t]
\renewcommand{\arraystretch}{1.5}
\caption{The properties of distributed machines learning techniques, where ``\checkm'' indicates that the technique has the corresponding
capability or functionality, and ``\cross'' indicates otherwise}
\centering
\begin{tabular}{|c|c|c|c|c|c|}
\hline

\diagbox{Techniques}{Features}&Dataset partition & Model partition  & Stored dataset  & Fast convergence &Optimal solution  \\ \hline
Federated learning  &\checkm    &\cross   &\checkm    &\cross    &\checkm  \\ \hline
Federated reinforcement learning  &\checkm     &\cross      &\cross    &\cross    &\checkm \\ \hline
Federated deep learning         &\checkm     &\cross     &\checkm    &\cross   &\checkm    \\ \hline
Partitioned learning  &\checkm   &\checkm    &\checkm   &\cross    &\checkm     \\ \hline
\end{tabular}
\label{tab.feat}
\end{table*}

\section{Convergence and Divergence of DML Algorithms} \label{sec.converge}
In general, a learning algorithm performs better with more iterations,
which, on the other hand, results in higher computational complexity.
In this section, we analyze the performances of popular and novel algorithms applied to DML
in terms of convergence speed, optimality, and the mathematical gap between the output and  labels of input data,
as summarized in Table \ref{tab.converge}.
The convergence of an algorithm indicates that the sequence of solutions can approach
a solution point within an infinitely small radius.
Divergence measures or approximates the gap between the probability distributions of the sample set (input data)
and the output data.

\begin{table*}[t]
\renewcommand{\arraystretch}{1.5}
\caption{A summary of popular and novel algorithms for DML}
\centering
\begin{tabular}{| p{2.5cm}<{\centering} | p{12cm}<{\centering} |}
\hline
Algorithm &Key features  \\ \hline
ADMM \cite{iut16, hong17}  &Linear scaling, linear convergence in some cases, preventing the disappearance of the gradient,
not sensitive to poorly-qualified input data     \\ \hline
DL-ADMM \cite{dladmm19} &Global convergence for DNN with a complexity of $\mathcal O(N^2)$,
updating parameters backward-forward    \\ \hline
Communication efficient-FedAvg \cite{jed20} &Sparsifying gradients and quantizing the weight into $8$-bit unsigned integers,
convergence at the required precision with six times fewer rounds than FedAvg, and three times less data transmission \\ \hline
Momentum gradient descent \cite{liu19mgd}     &Global convergence and prominent convergence enhancement over
first-order gradient descent       \\ \hline
Parallel mini-batch SGD \cite{dikel12} &Convergence speed of $\mathcal O(1/\sqrt{NT})$, $N$ times faster than the single-node SGD,
requiring simultaneous local updates from all nodes per iteration  \\ \hline
Decentralized parallel SGD \cite{aji17, alistar17, jiang18}   &Convergence speed of $\mathcal O(1/\sqrt{NT})$,
global gradient aggregations in PMSGD substituted by local collections between neighbors \\ \hline

\end{tabular}
\label{tab.converge}
\end{table*}

\subsection{Convergence of Deep Learning}

SGD and its variants are popular techniques to train DL models.
Yet, SGD is subject to multiple constraints.
For instance, the inaccurate data dwindles as the gradient is backpropagated (i.e., the vanishing of the gradient),
and the SGD gradient can be influenced significantly by a small portion of poorly-qualified admitted data.
On the other hand, ADMM decomposes a problem into several parallel subproblems and orchestrates overall scheduling across them to solve the original problem.
The merits of ADMM are:
i) it demonstrates linear scaling as data is examined and analyzed in parallel across units;
ii) it does not demand gradient update steps and prevents the disappearance of the gradient;
and iii) it is not sensitive to poorly-qualified input data.

The linear convergence speed of ADMM is evaluated in~\cite{iut16} under the assumption that the optimal solution to the problem
is achieved at the point $x^*$.
The objective function $f$ is assumed to be twice-differentiable at $x^*$ with $\nabla f(x^*)>0$.
Hong and Luo \cite{hong17} construct the global root-linear convergence measure of the ADMM, 
supposing that a predefined accuracy requirement is satisfied and the step size is small enough.
This accuracy requirement estimates the Euclidean distance from any iteration step to the best solution set with respect to the remaining neighborhood.
The accuracy requirement is satisfied, if the feasible set is a compact polyhedron while the target is a combination of
a smooth and strongly convex function and a non-smooth $\ell_1$ regularizer.
ADMM has linear convergence  when employed in many practical scenarios,
for instance, in the LASSO (Least Absolute Shrinkage and Selection Operator) method used for linear regression,  without requiring strict convexity of the target function.

Yet, the ADMM does not always guarantee fast and global convergence.
To tackle this shortcoming, Wang \emph{et al.} \cite{dladmm19} propose a deep learning ADMM (DLADMM),
which refreshes key parameters first backwards and then forwards.
It avoids the operation of matrix inversion by implementing the quadratic approximation and backtracking methodologies,
diminishing the computational complexity from $\mathcal O(N^3)$ to $\mathcal O(N^2)$.
Global convergence is proven for an ADMM-based DNN.

The momentum-based variants of SGD have been the leading techniques for solving ML problems.
For large-scale decentralized ML problems, such as training DNNs, a parallelized realization of SGD,
namely, parallel mini-batch SGD (PMSGD), is utilized in \cite{dikel12}.
With $N$ parallel workers, PMSGD has a convergence speed of $\mathcal O(1/\sqrt{NT})$,
$N$ times faster (referred to as ``linear speedup'') than the $\mathcal O(1/\sqrt{T})%
$\footnote{If an algorithm has $\mathcal O(1/\sqrt{T})$ convergence,
then it takes $1/\varepsilon^2$ iterations to achieve an $\mathcal O(\varepsilon)$ precise solution.}
convergence achieved by SGD at a single node \cite{lan13}.
Such linear acceleration with respect to the number of participating agents is desirable in decentralized learning when the
 number of active agents is large.
However, the linear acceleration is generally not easy to achieve, since the classic PMSGD demands all participating agents to update their local gradients or models simultaneously at every iteration.
Inter-node communication cost imposes a serious challenge~\cite{yu19linear}.

To rule out potential communication impediments, for instance, high delay or poor bandwidths,
multiple decentralized SGD variants have been developed.
For instance, decentralized parallel SGD is considered in \cite{assran18}, where global gradient aggregations
employed in the classic PMSGD are substituted by local communications and collections between neighbors.
To decrease the data transmission in each round, compression- and sparsification-based parallel SGD methods
are investigated in~\cite{aji17, alistar17}.
It is proven in \cite{jiang18} that certain parallel SGD variants that strategically skip communication rounds can provide the fast
$\mathcal O(1/\sqrt{NT})$ convergence with substantially fewer transmission rounds.
Recently, the momentum methods have been increasingly utilized in learning ML models and can usually converge more quickly
and be extended to more scenarios.
For instance, decentralized SGD with momentum is utilized for training DNNs with big data.

\subsection{Convergence of Federated Learning}
Federated Averaging (FedAvg) is a popular FL algorithm, which suffers from a slow convergence with large numbers of rounds and high communication costs per round under non-identical and independent distributed (non-i.i.d.) client datasets.
Mills \emph{et al.} \cite{jed20} propose an adaptive FedAvg technique.
The technique leverages a decentralized realization of the Adam optimization~\cite{adam14}, to reduce the number of rounds significantly before convergence.
A new compression technique is developed to reduce the transmission load of non-i.i.d. datasets by sparsifying gradients and
quantizing the weight into $8$-bit unsigned integers, which results in Communication-Efficient FedAvg (CE-FedAvg).
Numerical tests are carried out using the MNIST/CIFAR-10 datasets, identical and independently distributed (i.i.d.) or non-i.i.d. data,
different numbers of engaging clients, client participation ratios, and different data-reduction schemes.
It is demonstrated that the CE-FedAvg is able to stabilize at the required precision with six times fewer rounds than its FedAvg counterpart.
Meanwhile, CE-FedAvg transmits three times less information.
Field trials with Raspberry-Pi devices validate that CE-FedAvg can reach a required precision at most $1.7$ times faster than FedAvg.

The decentralized FedAvg algorithm is proposed to protect data privacy in \cite{mahan17} and \cite{Wu2019Distributed},
which combines the local SGD of each wireless participant with a server that carries out model averaging.
A number of edge nodes are randomly and non-repetitively selected every round to
synchronize their local models with the global model and restart their respective training processes.
A decentralized circular topology is adopted in \cite{Wu2019Distributed}, where there exists at least one path between two nodes (or vertices).
Extensive experiments are run on the FedAvg algorithm under non-i.i.d data from the MNIST training dataset.

The existing studies on FL solely leverage the first-order gradient descent (GD), without taking into account past iterations
for the gradient renewal, which can potentially speed up convergence.
Liu \emph{et al.} \cite{liu19mgd} consider a momentum element which depends on the previous iteration.
A momentum FL (MFL) technique employs the momentum gradient descent (MGD) to update the model parameters
of different workers or agents.
The authors analyze the global convergence behaviors of the MFL and establish an upper limit for the convergence speed.
By comparing  the upper limits of the MFL and FL, conditions are provided under which the MFL can achieve faster convergence. For different ML frameworks, the convergence behavior of MFL is examined experimentally using the MNIST dataset.
Simulation results justify that MFL can converge globally and provides a prominent convergence enhancement over FL.

\subsection{Divergence}
The Kullback-Leibler (KL) divergence is a popular way to measure the mathematical gap between two different probability distributions, i.e., $p(\cdot)$ and $q(\cdot)$~\cite{rached04, boche06, pinski15, bu18}.
The entropy $H(p)$ of a distribution $p$ describes the minimum possible number of bits per message that are needed (on average)
to successfully encode events drawn from $p$.
The cross-entropy $H(p,q)$ indicates the number of bits per piece of information needed (on average) to encode events drawn from the true distribution $p$, if using an optimal code for distribution $q$~\cite{rao85, miller93, boer05}.
Given $p$, $H(p,q)$ grows as $q$ becomes increasingly different from $p$.
However, if $p$ is not fixed, $H(p,q)$ does not provide an absolute measure of the difference. The reason is that $H(p,q)$ grows with the entropy of $p$.
Given the distribution $p$, the KL divergence and the cross-entropy are interchangeable.
As revealed in~\cite{jones90, hino11, frank16},
when $p$ and $q$ are identical, the cross entropy is non-zero and equal to the entropy of $p$.

Cross-entropy is commonly used as part of loss functions in ML, including FL~\cite{sanga16, tok19}.
In many cases, $p$ is treated as the ``true'' distribution, and $q$ as the model to be optimized~\cite{hadar19}.
For example, in categorization tasks, the commonly used cross-entropy loss (also known as log-loss),
quantifies the cross-entropy between an empirical distribution of the features (e.g., by fixing the admitted data samples)
and the distribution estimated by the classifier~\cite{rao85}.
For each data point, the empirical distribution assigns probability $1$ to this data (class), and probability $0$ to all other classes.
In this example, the cross-entropy changes proportionally with the negative log-likelihood.
As a result, minimizing the cross-entropy amounts to maximizing the log-likelihood.
Given $p$ (i.e., the empirical distribution in this example),
minimizing the cross-entropy amounts to obtaining the smallest KL divergence value between the empirical distribution
and the predicted.
$H(p)$ is not influenced by the model parameters and is negligible in the loss function.

The KL divergence and cross-entropy have been used to optimize wireless systems \cite{palang16, sun16},
e.g., by assuming mutually dependent sparse Multiple Measurement Vectors (MMVs) with unknown dependency.
Palangi \emph{et al.} \cite{palang16} characterize this reliance by calculating the conditional probability of each non-zero element in a vector,
with knowledge of the ``remainings'' of all previous vectors.
To estimate the probabilities, the long short-term memory (LSTM) is utilized to develop a data-driven framework for sequence modeling.
A cross-entropy cost function is minimized to learn the model parameters.
To re-establish the sparse vectors at the decoder, a greedy approach is developed to use the model to decide tentatively the conditional probabilities in~\cite{palang16}.
Numerical experiments on two practical datasets validate that the developed approach substantially outperforms
a general-purpose MMV approach (i.e., the Simultaneous Orthogonal Matching Pursuit) and several model-based Bayesian schemes.
The approach developed in~\cite{palang16, sun16} does not append any intricacy to the compressive sensing encoder.

Geographically widespread systems in an end-edge-cloud structure are popularly employed
in healthcare systems~\cite{endwang19, lyuxc20},
and can be potentially applied to wireless network environments for communications among IoT devices~\cite{endren19}.
FL is valuable for such practical scenarios. 
The convergence of FL has been widely studied in ubiquitous
systems~\cite{endzeni19} and many results are exportable to the FL implementation over wireless networks.
Generally speaking, existing FL schemes cannot cope with inequalities of local data features.
Yet, the agents in ubiquitous systems are subject to label noises because of different capabilities, biases,
or malicious interferences of the annotators.
Chen \emph{et al.} \cite{focus20} propose Federated Opportunistic Computing for Ubiquitous Systems (FOCUS) to tackle the label noises.
FOCUS keeps a small group of baseline samples at the server, and quantizes the accuracy of the agent local data by calculating the joint cross-entropy between the FL model on the local sample data and the FL model on the baseline sample data.
Then, an agent's weighted operation is carried out to calibrate the weight allocated to the agents in the FL model according to their accuracy scores.
FOCUS is tested by experiments by using both synthetic data and practical data.
The outcomes demonstrate that FOCUS distinguishes the agents with noisy labels and diminishes their influence on the model performance,
hence scoring substantially higher than the FedAvg approach.

\section{Computation and Communication Efficiency in DML}

A large dataset and number of communication rounds can ensure a better training accuracy.
Yet, the massive data trained and transmitted in the system incurs high dimensionality, complexity,
and computation and communication cost.
In this section, we discuss the approaches to improve computation and communication efficiency,
and achieve a balance between computation and communication,
with a guaranteed model performance.

\subsection{Computation}

Large-scale datasets in ML systems invoke computational challenges,
requiring distributed training on an aggregated group installed with accelerators, such as GPUs~\cite{you17}.
To train a deep model in decentralized networks, typical mathematical operating missions are performed by the GPUs of multiple workers in numerous iterative loops~\cite{kim17}.
Every iterative loop consists of a few key stages, namely, feed-forward and back-propagation.
During every iteration, those GPUs perform massive data transmissions about the model parameters~\cite{wang17arx}
or gradients~\cite{raina09}.
Therefore, information transmission across GPUs is a potential impasse impeding the system-wide learning effect.
Shi \emph{et al.} \cite{shi18} develop a model based on directed acyclic graph (DAG), which is widely used for distributed (or federated) DL
to reduce the training time based on the decentralized synchronous SGD technique.

A typical DML model is presented in~\cite{amiri19dec}, where computation tasks across $n$ workers (or agents) are coordinated for a large-scale decentralized learning problem facilitated by a master.
Superfluous calculations are allocated to the agents to prevent stragglers.
SGD is utilized for each computation round, which is considered finished as long as the master acquires $k$ distinguished calculations,
i.e., \emph{calculation objective}.
The \emph{average execution duration} is a function with respect to the \emph{calculation workload}
(given by the proportion of the data samples usable at every agent), as well as the calculation objective.
A lower limit of the shortest average execution duration is established with assumptions on the prior information of the random calculation and transmission latency.
Experiments performed on the Amazon EC2 cluster demonstrate considerable decrease in terms of the average execution duration, as compared to current coded and uncoded calculating approaches.
The difference between the developed approach and the lower limit is insignificant,
justifying the merit of the developed coordinating policy.

A typical DML model can be implemented with different off-the-shelf DL frameworks. Four of the latest such frameworks are Caffe-MPI from Inspur~\cite{Caffe-MPI}, CNTK from Microsoft~\cite{CNTK}, MXNet from Apache~\cite{mxnet} and TensorFlow from Google~\cite{Tensorflow}. In \cite{shi2018performance}, the four frameworks are tested, and it is shown that the different frameworks lead to different performances, even when the same GPUs and datasets are utilized. Running DNNs that efficiently learn the feature representation, the four frameworks are popular for AI applications. The large-scale datasets of DNNs may lead to a high requirement of computation resources. GPUs are utilized to accelerate training by both algorithmic parallelization and data parallelization. Implemented in GPUs, the four frameworks are tested and compared in terms of training speed in a single GPU or multiple GPUs. Experiments help identify bottlenecks preventing high throughput dense GPU clusters, and potential remedies, such as efficient data pre-processing, autotune and input data layout for loading massive data, and a better-utilized cuDNN (i.e., an accelerator library for DNN invented by NVIDIA \cite{cuDNN}).
In this sense, the joint optimization of the DML model and implementation frameworks is critical to the performance of the model.

\begin{figure*}
\centering
\includegraphics[width=0.7\textwidth]{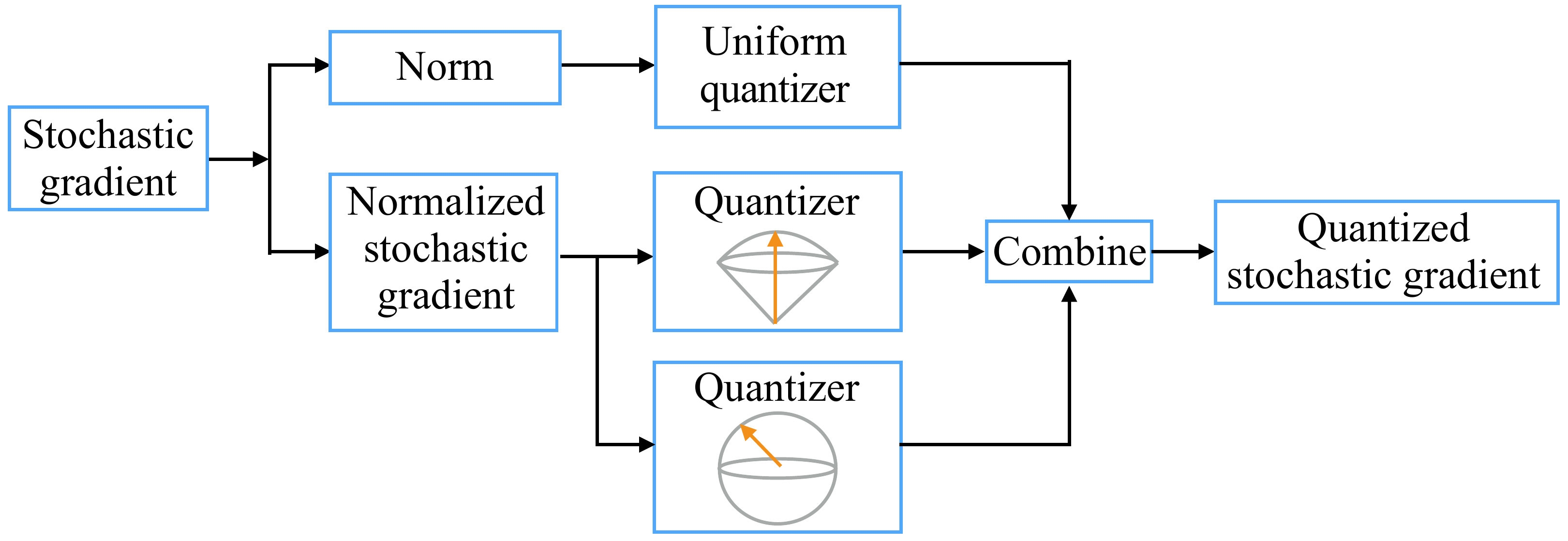}
\caption{A hierarchical gradient quantization scheme \cite{du19}.
The normalized block gradients are quantized with a uniform quantizer.
Then, the quantized block gradients are numerically adjusted and connected.
The size of the conjunction vector is reduced by another low-dimensional quantizer.
Featuring bit-assignment, the developed scheme reduces the measurement error
by dividing the overall number of bits from gradient measurement to decide on the simplifications of quantizers.}
\label{du19quan}
\end{figure*}

\subsection{Communication}
To utilize FL under a decentralized setting, the agents have to transfer the model parameters via wireless channels.
The wireless links can cause training inaccuracies because of scarce wireless resources (such as bandwidth) and lossy wireless links~\cite{chen20}.
For instance, the symbol errors caused by the lossy wireless channels and limited wireless resources
can have a strong influence on the efficiency and precision of the FL renewals among the agents.
The inaccuracies influence the effect and convergence rate of the FL algorithms~\cite{jakub16ce}.

Federated optimization is proposed in \cite{jakub16} to train a centralized model,
meanwhile keeping the training parameters locally at users' devices.
The purpose is to diminish uplink transmission costs and protect the data privacy of the devices with an emphasis on
sparse data, where some attributes take place on a small subset of nodes or data points only.
It is shown that the sparsity characteristic can be utilized to derive an effective algorithm for federated optimization,
which can reduce communication rounds and network bandwidth required by model training.
The federated optimization would still benefit from further improvements,
such as asynchronous model uploading, to achieve better performance.

As a popular edge learning technique, FL has been developed according to decentralized gradient descent,
where the stochastic gradients are calculated at the agents situated in the edge and sent to the server also located in the edge
to update an overall ML model.
Because each stochastic gradient consists of millions to billions of parameters,
information exchange overhead turns out to be a potential impasse of edge learning.
To circumvent the impasse, Du \emph{et al.} \cite{du19} develop a graded architecture to decompose and measure stochastic gradient
(as illustrated in Fig. \ref{du19quan}) and evaluate its influence on the training effect.
The system decomposes the stochastic gradient into several normalized block gradients.
The normalized block gradients are quantized with a uniform quantizer.
Then, the quantized versions of the normalized block gradients are numerically adjusted and connected.
The quantized normalized gradient utilizes a so-called conjunction vector for the least distortion.
The size of the conjunction vector is also effectively reduced by another low-dimensional quantizer.
Featuring bit-assignment, the developed scheme reduces the measuring mistake
by dividing the overall number of bits from gradient measurement to decide the simplifications of quantizers.
The scheme is shown to ensure the convergency by establishing the convergence speed to be a function of the number of bits used for the uniform quantizer.
Numerical tests validate that this architecture can substantially diminish the information exchange overhead,
in contrast to the sign-SGD method, while both yield similar training precisions.

While existing works try to reduce the total number of bits sent at every renewal by means of an effective reduction of the data size,
Wang \emph{et al.} \cite{CMFL19} investigate an orthogonal method that distinguishes impertinent renewals provided by clients and prevents them from being transmitted to diminish system footprint.
Communication-Mitigated Federated Learning (CMFL) is developed in~\cite{CMFL19}, which provides the clients with the feedback of the
\emph{global inclination} of model refreshing.
The global inclination is evaluated by the divergence between the gradients of a local renewal and a global renewal.
Each agent checks whether its renewal is consistent with this overall inclination and also sufficiently pertinent to any model enhancement.
By preventing transmission of misaligned renewals to the server, CMFL is able to significantly reduce the transmission cost
and ensure the training convergence at a given prediction accuracy.
CMFL can enhance the transmission efficiency in comparison to many existing FL schemes,
such as vanilla FL \cite{mahan17}, Gaia \cite{hsieh17}, and Federated Multi-Task Learning (FMTL)~\cite{smith17}.
CMFL is evaluated by numerous simulations and EC2 emulations~\cite{ec2}.
In contrast to vanilla FL, CMFL achieves $13.97$ times more transmission efficiency,
while Gaia only yields $1.26$ times enhancement in efficiency.
When utilized to FMTL, CMFL enhances the transmission efficiency by $5.7$ times with $4\%$ better estimation precision.

\subsection{Trade-off Between Computation and Communication}
Emerging technologies and applications, such as IoT and social networking, obtain services from a large group of people and devices,
and produce massive data at the system edge.
ML models are generally constructed from aggregated information, to allow the observation, categorization, and estimation of future events.
With practical considerations on bandwidth, data plan, storage and privacy of users or devices,
it is usually infeasible to upload all information to a central server.
Wang \emph{et al.} \cite{wang19} focus on the FL models trained with gradient descent-based algorithms.
The convergence limit of decentralized gradient descent is analyzed under non-i.i.d data,
based on which a control protocol is developed to balance the local renewal
and the global data collection, and obtain the smallest loss function constrained by a predetermined budget.
The effect of the approach is validated experimentally with sample data,
both on a network prototype and a large-size simulation system.
The experimental results demonstrate that the algorithm behaves near-optimally in terms of the global loss function
with a range of ML models and different data distributions.

Unlike the conventional DAG model whose nodes denote calculating missions,
the DAG model developed in \cite{shi18} consists of two groups of nodes: calculation and transmission nodes.
To examine the influence of information exchange on the learning effect,
Shi \emph{et al.} \cite{shi18} carry out empirical investigations on four classic decentralized DL architectures
(including Caffe-MPI, CNTK, MXNet, TensorFlow, etc.) and test a range of information exchange policies,
such as PCIe, NVLink, 10GbE, and InfiniBand.
The speedup of training typically relies on three factors: Input/Output, calculating, and transmission behaviors.
By running synchronous-SGD with multiple GPUs, these DL architectures can decouple the collection and calculation of the gradients into parallel operations to accelerate model training.
The frameworks do not scale well using some latest GPU products (e.g., NVIDIA Tesla V100).
The reason is that the present integrations of inter-node gradient transmission through $100$ Gbps InfiniBand
are not fast enough to keep up with the computation capability of V100.

When data is distributed across multiple servers, it is critical to lower the information exchange cost between the servers
(or participating agents) while resolving this decentralized learning task.
Elgabli \emph{et al.} \cite{anis20} develop a quick and transmission-efficient distributed algorithm to tackle the decentralized ML task.
Developed from ADMM, this algorithm of Group ADMM (GADMM) solves the problem
in a distributed fashion, where no more than half of the agents compete for the scarce radio resources at a time.
In particular, the agents are separated into two clusters (namely, \emph{head} and \emph{tail}).
Every agent in the head (tail) shares data only with its two neighbors from the same cluster,
hence training an entire model with small communication cost in each model exchange.
It is proved that GADMM converges to the optimum, as long as its loss functions are convex.
It is experimentally corroborated that GADMM converges more quickly and is more transmission-efficient than classic methods,
for example, Lazily Aggregated Gradient (LAG)~\cite{lagc20},
when linear or logistic regression problems are created based on synthetic and practical data.
Additionally, dynamic GADMM is developed to achieve convergence when the network topology of agents changes over time.



\section{DML Architectures}




Several popular architectures have been widely adopted in DML, and often applied to the studies of wireless communication networks.
Table IV summarizes their properties and features with details provided in the following.

\subsection{MapReduce}
\begin{figure*}
	\centering
	\includegraphics[width=0.75\textwidth]{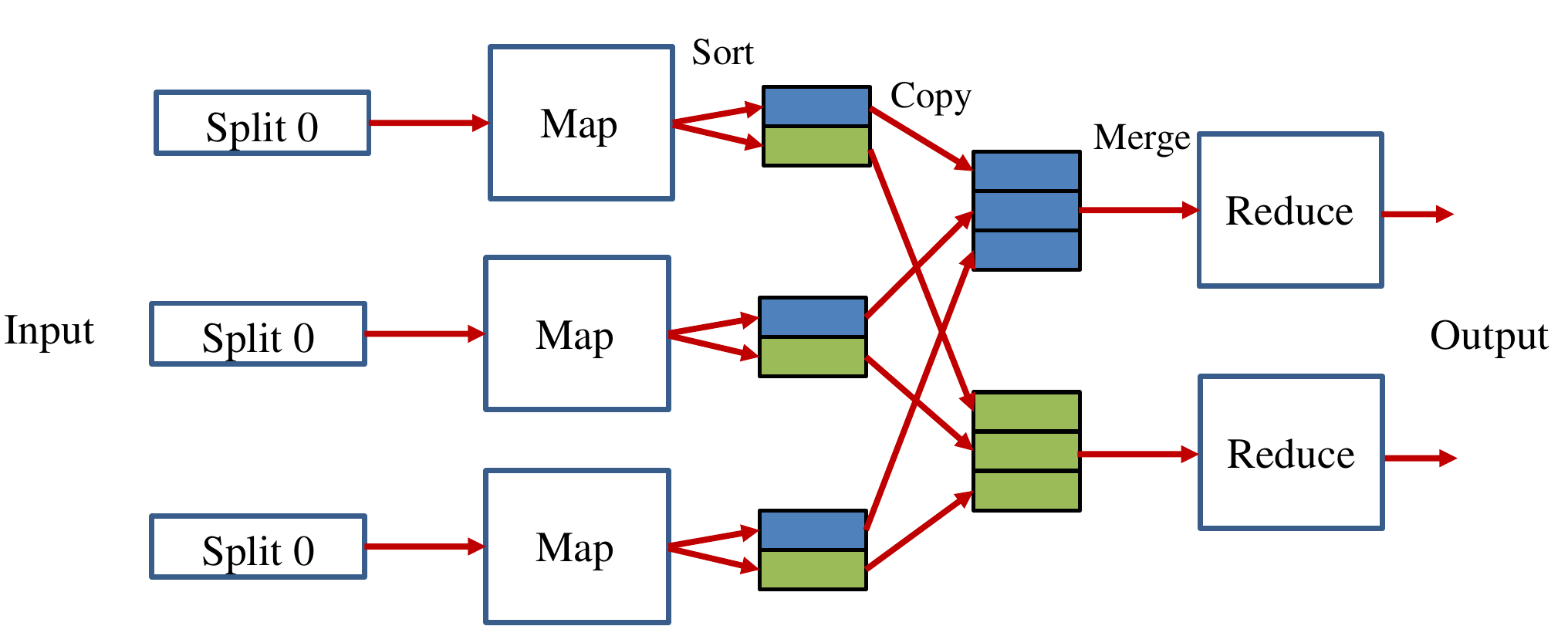}
	\caption{An illustration of the MapReduce architecture consisting of a Map and a Reduce function. The Map function integrates the intermediate values corresponding to a same intermediate key. The Reduce function merges the input values from the map to yield a smaller number of values  \cite{White12Hadoop} .}
	\label{Mapreduce}
\end{figure*}
MapReduce is a parallel programming paradigm to process and generate large datasets, which is applicable to various ML tasks in practice. Users complete the computation by map and reduce operations. The system decentralizes the computing tasks and runs the tasks in parallel over clusters of machines, and manages communications and synchronization between the machines for efficient usage of the network and memories~\cite{dean2008mapreduce, hou20, hou2020}.
Through MapReduce, it is easier to parallelize plenty of batch data computing tasks. It is suitable for large-scale clusters without considering failover management. With emerging open-source platforms (e.g., Hadoop), MapReduce can accelerate real-world applications, e.g., web browsing and fraud detection.

Fig.~\ref{Mapreduce} illustrates the data flow of MapReduce\cite{White12Hadoop}, which is comprised of a Map and a Reduce function.
The role of the Map function is to integrate the intermediate values corresponding to the same intermediate key, which are then transferred to the Reduce operation. Map generates some intermediate key and value pairs from an input pair. The Reduce function, in contrast, merges the input values to yield a smaller number (typically one or zero) of values. 
While MapReduce is highly scalable, it suffers from a crucial weakness for ML: It fails to recognize the iterative nature of most ML algorithms \cite{Rosen12IterativeMap}. To address this issue, ML systems based on iterative MapReduce (IMR) framework were proposed, including Spark MLlib, Vowpal Wabbit and Twister. The IMR-based ML systems employ asynchronous iterative communication pattern.

Several systems have been designed based on the MapReduce architecture on the Hadoop platform. Hadoop provides a framework for analyzing and transforming very large
datasets using the MapReduce paradigm.
Jeon \emph{et al.} \cite{SungHwan2018MapReduce} show how MapReduce
parameters can affect the distributed processing of ML programs, which are supported by ML libraries like Hadoop Mahout. Virtualized clusters are constructed on top of Docker containers to measure DML performance, while changing Hadoop parameters. It is shown that the processing becomes faster with the decreasing number of replica and with the increasing block and memory buffer sizes.
In \cite{astekin2018evaluation}, a decentralized realization of two unsupervised methods, i.e., $K$-means clustering and primary component access (PCA), are evaluated for robust anomaly detection method in a distributed system. The experiments are performed with Hadoop Distributed File System (HDFS) log dataset. The conclusion drawn is that the accuracy of the methods does not degrade because of the decentralization. Both $K$-means clustering and PCA can detect 97\% of actual anomalies. When the parallelization level is high, both systems have faster running speeds (e.g., 30\% faster when parallel threads grow from 1 to 10),
and the efficiency of anomaly detection increases by 75\%.

More systems have been designed based on the MapReduce architecture on the Spark platform.
Patil \emph{et al.} \cite{Patil2019accuracy} propose an implementation of Distributed Decision Tree (DDT) in Spark environment, which can reduce the model building time without compromising the accuracy of the Decision Tree (DT).
Apache Spark clusters provide a distributed experimental environment to validate DDT training. Three models are considered in \cite{Patil2019accuracy}: Spark Decision Tree (SDT), PySpark, and MLlib. It is shown that, when the size of dataset is up to gigabytes, the traditional DT (e.g., PySpark) is outperformed by DDT in terms of learning time (e.g., by 8 minutes on average). Although the SDT is similar to MLlib in terms of runtime, one can have a Python wrapper around SDT which would not restrict the input data format, speeding up the training. For this reason, the SDT is recommended in \cite{Patil2019accuracy}.

A scalable system for heart disease monitoring based on Spark and Cassandra frameworks is described in \cite{ed2019real}. This system focuses on the real-time classification of heart disease attributes. The system consists of two main subsystems, namely, streaming processing, and data storage and visualization. The former uses Spark MLlib with Spark streaming and classification models to forecast heart disease. The latter uses Apache Cassandra to store the large volume of generated data. 
An architecture is built for a recommender system in \cite{sewak2016IoT},
which recommends products or services to consumers, relying on their optimal responses to the recommendations.
DML algorithms (e.g., variants of Decentralized Kalman Filters, Decentralized Alternating Least Square Recommenders, and Decentralized
Mini-Batch SGD-based Classifiers), and calculation and ML platforms with strong scalability (e.g., Apache Spark, Spark MLlib, Spark Streaming, and Python/PySpark) are integrated to build a high-performance, decentralized, and fault-tolerant architecture.

Dong \emph{et al.} \cite{dong2018forecasting} develop a forecasting model to predict future electricity cost of residents via a Random Forest ML algorithm.
Distributed systems, such as Amazon Web Service (AWS), Simple Storage Service (S3), Elastic Map Reduce (EMR),
MongoDB and Apache Spark, are used to store and process residential energy usage data collected by a smart meter in London.
A significant computational gain of using decentralized networks when implementing ML
algorithms on large-scale data is observed in terms of computational times and predictive accuracy.

\subsection{Parameter Server (PS)}
\label{sec.ps}
\begin{figure}
	\centering
	\includegraphics[width=0.38\textwidth]{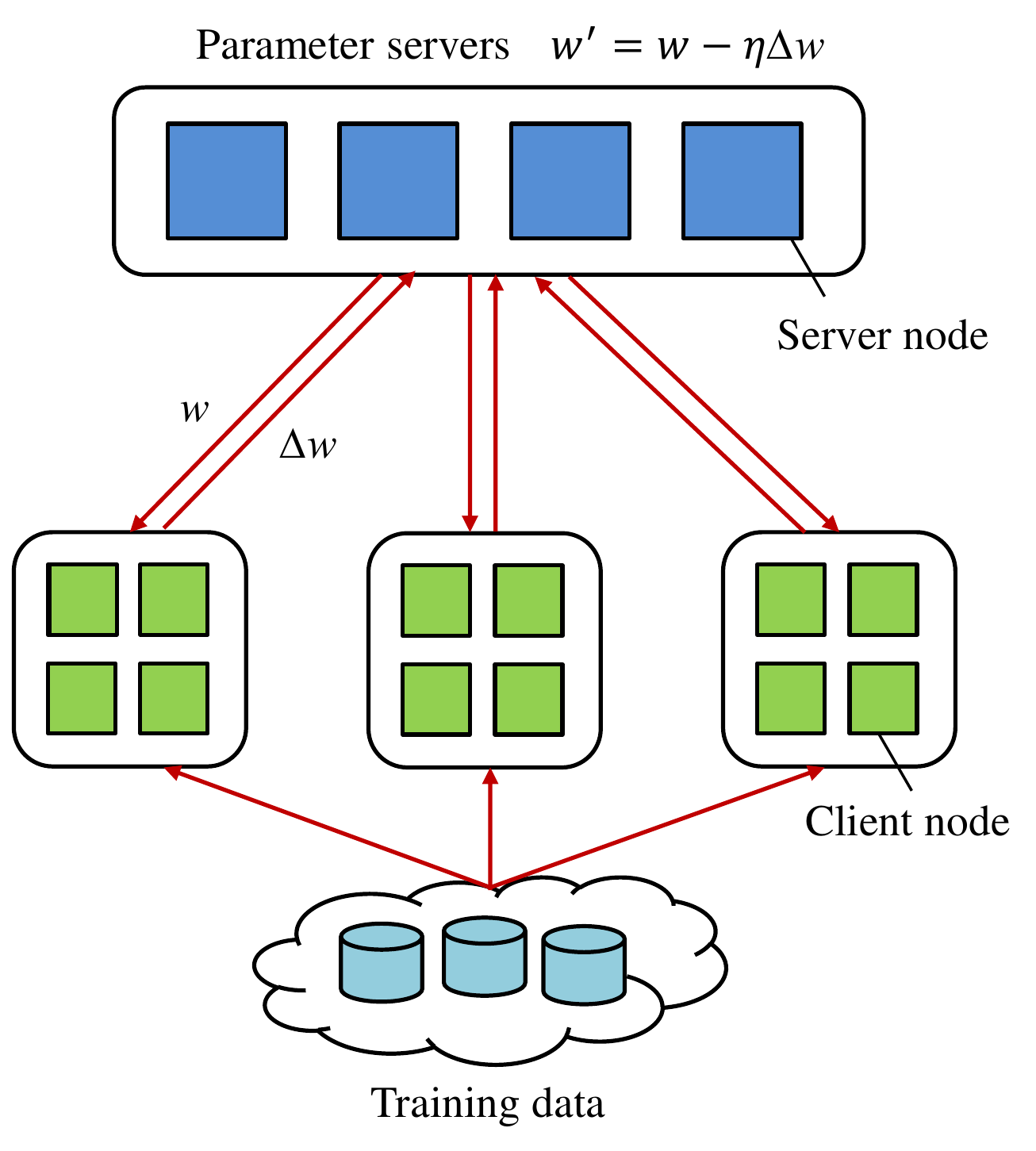}
	\caption{An illustration of the parameter server architecture. The client nodes are partitioned
into groups. The servers maintain the whole or part of all parameters and aggregate the weights from each client group.}
	\label{PS}
\end{figure}
The PS architecture consists of two kinds of nodes: server and client (or worker); see Fig.~\ref{PS}. There may be one or multiple servers. The client nodes are partitioned into groups. The servers maintain the whole or part of all parameters, and aggregate the weights from each client group. The server nodes interact with each other to duplicate and transfer parameters \cite{limu13}.
In most cases, the server nodes and the hosts in the cluster are of equal numbers. Every individual server node only needs to synchronize partially with each other. This reduces data transferred between the server and client \cite{zhang2018quick}.

The client nodes conduct the initial steps of the DL algorithm, such as convolutional calculation, gradient calculation, back propagation, weight refreshment. Unlike a centralized approach, a client uses the synchronized global gradient from the server nodes to carry out back propagation and weight refreshments. The clients only share the parameters with the servers, and never communicate with each other. They typically store locally part of the training data \cite{limu14}. Communications between nodes in a parameter server architecture can be asynchronous. By relaxing the stringent requirement of synchronization, the efficiency of the system can be improved by parallelizing the utilization of the CPU, disk and network bandwidth \cite{limu13}. 
The PS architecture has been broadly applied to decentralize ML tasks on wired platforms.

Given the finite bandwidth and memory, and potential compromise of privacy, it is generally
impossible to deliver all data in a distributed edge environment.
In \cite{wang2018edge}, an adaptive control technique is developed to strike the balance of learning efficiency and wasted resources in real-time, by specifying the frequency of the global aggregations.
The technique minimizes the learning loss to adapt to the frequency of the global aggregations.
Non-i.i.d. datasets are considered for learning tasks. The tradeoff between the efficiency of the global and local updates is improved by calibrating the time loss, the number of active nodes, and the sensitivity of fixed control parameter under various data distributions and node numbers.
In \cite{Tuor2018Demo}, the authors of \cite{wang2018edge} demonstrate the system developed in \cite{wang2018edge}. It is shown that the system can make an estimation of the parameters with regard to data distribution and resource usage, and adjust the system relying on the estimations on-the-fly.

Duong and Sang \cite{Duong2018Distributed} propose and implement FC$^2$: A web service for fast, convenient and cost-effective (FC$^2$) DML model training over the cloud. The designed system can exploit inherent heterogeneity in computing resources over the cloud, and provide a simple web-based interface for cost-effective training of DML models. 

\subsection{Graph Processing Architecture}

\begin{table*}[t]
\renewcommand{\arraystretch}{1.5}
\caption{Features of DML architectures}\label{feautures-arch}
\centering
\begin{tabular}{|c|c|c|c|c|}
\hline
  \diagbox{Architectures}{Features}   & Communication pattern & Parallel type& Flexibility & User friendliness\\ \hline
  Iterative MapReduce  & Synchronous  & Data parallel  & Low & Well-developed system\\ \hline
  Parameter Server  & Synchronous/Asynchronous  & Data parallel  & High &None built-in ML algorithm \\ \hline
  Graph Processing  & Synchronous/Asynchronous  &  Data/Model parallel & Medium &-- \\ \hline

\end{tabular}
\label{tab.arch}
\end{table*}

%


\begin{figure*}
	\centering
	\includegraphics[width=0.8\textwidth]{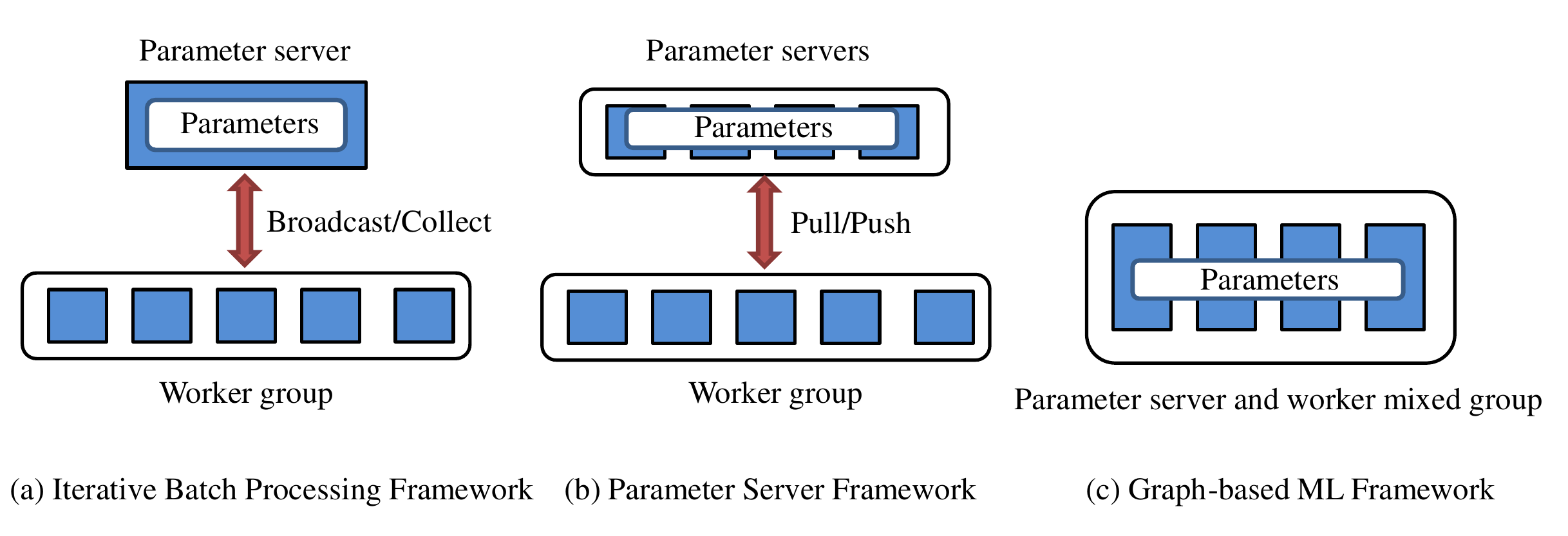}
	\caption{Comparison among DML frameworks. Compared with the other two frameworks, graph-based ML framework distributes both training data and model parameters within the same cluster, achieving good computing locality and reduced communication overhead \cite{Tian2018Cymbalo}.}
	\label{Cymbalo}
\end{figure*}

While other DML frameworks (such as iterative MapReduce and PS) only support data parallelism, recent graph-based frameworks consider model parallelism~\cite{low2012distributed,Tian2018Cymbalo,xiao2017tux2,chen2014bipartite}.
A comparative study of representative architectures is provided in Fig.~\ref{Cymbalo} \cite{Tian2018Cymbalo}.
Iterative batch processing frameworks, a.k.a. IMR, store their model parameters in
one node, broadcast to the workers and receive updates
from them during every iterative loop. The parameter server architectures store training data and parameters in two different groups of machines, and pull/push the parameters between each other. In contrast, the graph-based architectures treat the parameters as different kinds of vertices.
Distributed graph processing systems generally obey the ``think like a vertex'' idea, which allows a user-centric vertex function to execute at different vertices of a graph simultaneously. During every iterative loop, any vertex $v \in V$ is stimulated by its neighboring vertices along edges $e \in E$, or the system to compute, and votes to halt when completing the computation. A computing task terminates once all vertices vote to halt.

Early works on graph processing platforms (e.g., GraphLab \cite{low2012distributed}) was in light of ML, driven by the finding that a lot of ML problems have the potential to be formulated with graphs and tackled with iterative convergence techniques.
TuX$^2$ unifies graph and the parameter server framework, and creates a new framework \cite{xiao2017tux2}. The new framework applies
graph engines to the data representation and modeling, the programming model, as well as the operation schedule.
TuX$^2$ provides a programming model with edge-centric and vertex-centric interfaces. The edge-centric interfaces are designed to perform data exchange, while the vertex-centric interfaces are designed to perform update operations. However, many ML algorithms can be vectorized and translated to vector/matrix operations. When implementing an ML algorithm on TuX$^2$, it is complicated to translate vector evaluations to be graph-based over vertices and edges. A vector-centric model can avoid low-level implementation details and simplify the programming of ML algorithms.
Motivated by this, Tian \emph{et al.} \cite{Tian2018Cymbalo} propose an efficient graph processing platform to implement DML. The platform is called Cymbalo and has: i) data storage structure for heterogeneous data and a hybrid framework for ML operations, and ii) a vector-centric programming architecture that is more efficient than existing graph processing frameworks (e.g., BiGraph \cite{chen2014bipartite} and TuX$^2$). Simulations show that Cymbalo implemented on Spark 2.1.2 outperforms a lot of other advanced DML systems (such as Spark, PowerGraph,
and Angel) by expediting the ML operations by 1.6 to 5.8 times.

Distributed sensor networks consisting of identical and independent active nodes are deployed to provide the reliability and versatility in security applications. The networks have a strong resistance to network errors. 
In \cite{Picus2008Boltzmann}, a distributed sensor network is developed based on Boltzmann machine topology, where every node is modeled as a Boltzmann machine. The nodes in a sensor network can be naturally regarded as vertices in a graph processing framework. The network consists of sensors and inference units. The inference units detect information (e.g., videos) and decide whether to transmit to the next node or not. In the simulation presented in \cite{Picus2008Boltzmann}, hundreds of trajectories are produced to provide the inputs to the algorithm. It is found that the accuracy of the inference depends on the observations and positions. 



\section{Software Platforms and Limitations}
\subsection{MapReduce-Based Software Platform}
\begin{table*}[t]
\renewcommand{\arraystretch}{1.5}
\caption{Software platform and limitation, where ``\checkm'' indicates that the software platform has the corresponding limitation, and ``\cross'' indicates otherwise}
\centering
\begin{tabular}{|c|c|c|c|p{4cm}<{\centering}|}
\hline

  \multicolumn{2}{|c|}{\diagbox{Software platforms}{Limitations}} &Strict synchronisation  & Limited scalability & Others\\ \hline
  \multirow{5}*{Iterative MapReduce}
 &Hadoop \cite{Shvachko2010The} &\checkm  &\checkm    &Batch processing  \\ \cline{2-5}
   ~&Apache Spark  &\checkm  &  -- &Slow scheduling \\\cline{2-5}
    ~&EASGD-based platform in \cite{Gu2018Parallelizing}&\checkm &  --  &--   \\ \cline{2-5}
     ~&S4 \cite{neumeyer2010s4} &\checkm  &   -- &Stream processing, static routing \\\cline{2-5}
    ~&BAIPAS \cite{Lee2017BAIPAS}&\checkm &  --  &  -- \\ \cline{2-5}
    ~&Coded platform in \cite{Lee2018Speeding}  &\checkm  & --  &-- \\\hline
  \multirow{2}*{Parameter Server}&Petuum \cite{xing2015petuum}&\cross  & --  &Offline resource adjustment, bad fault recovery ability  \\ \cline{2-5}
  ~&Shared-memory platform in \cite{Lim2017accelerating}&\checkm & --   &--   \\ \hline

   \multirow{2}*{Graph Processing}&GraphLab \cite{low2012distributed}&\cross  &  --  &--\\ \cline{2-5}
  ~&Pregel \cite{malewicz2010pregel}&\cross  & \cross  &Lack of theoretical performance analysis  \\ \hline

\end{tabular}
\label{tab.software}
\end{table*}

\begin{figure}
	\centering
	\includegraphics[width=0.5\textwidth]{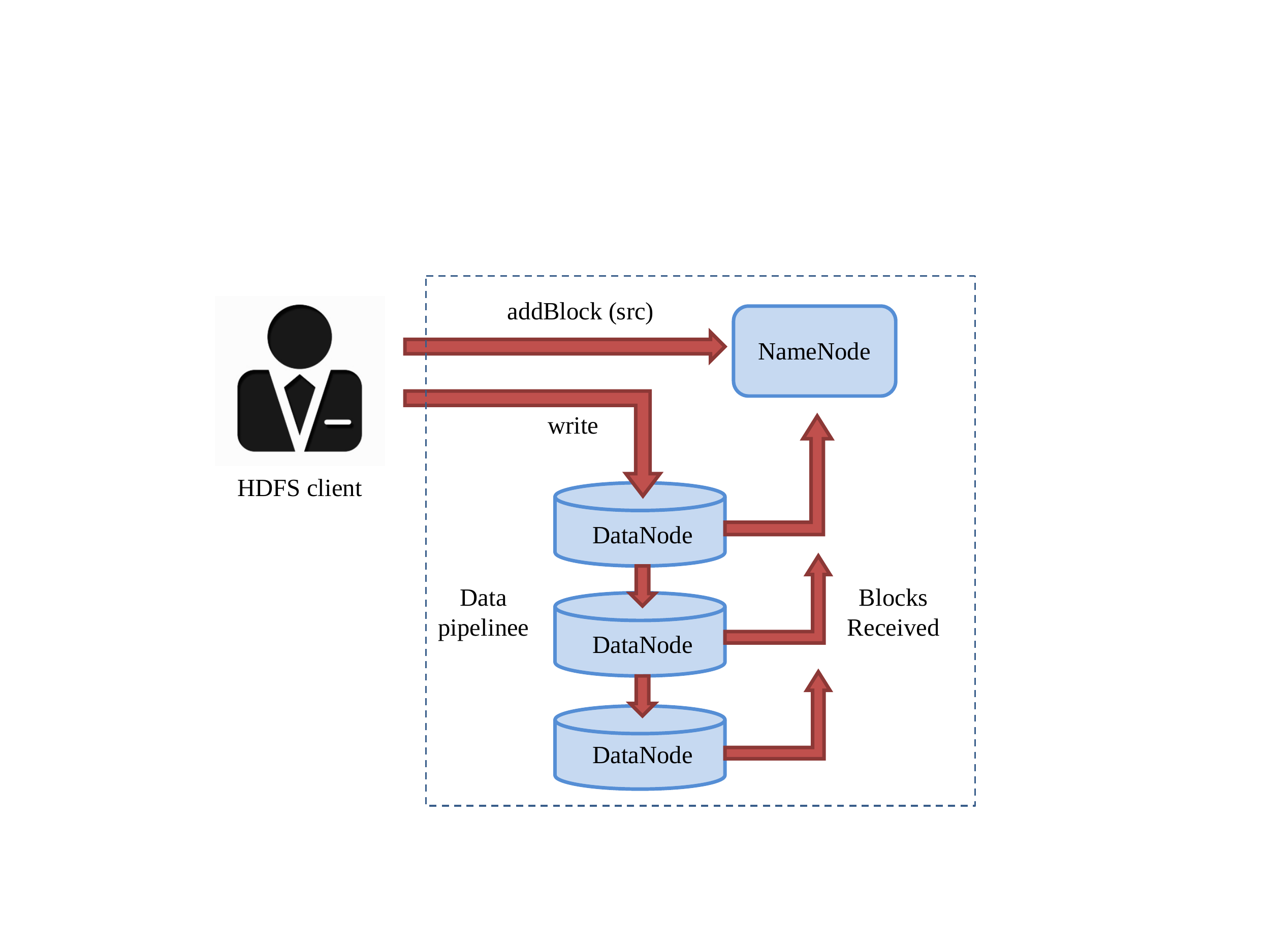}
	\caption{An HDFS system consisting of a HDFS client, a NameNode and several DataNodes \cite{Shvachko2010The}.}
	\label{HDFS}
\end{figure}

Hadoop is a MapReduce-based distributed file system to its partitioning of both the data and computing operations among a large number of hosts. Another property is that application
computations can be parallelized and executed at the point of capture. A Hadoop
cluster can readily upgrade its computing capability, storage volume, as well as the input/output (IO)
bandwidth, by plugging more off-the-shelf servers into the cluster.
In \cite{Shvachko2010The}, a Hadoop distributed files system (HDFS) is developed in an attempt to reliably accommodate huge datasets in servers and send the datasets to user/client applications. In the system, NameSpace is hierarchical storage for files and directories, which consists of multiple entities, namely, NameNode (specified by inodes and their memory attributes, such as permissions, modification and access times) and DataNode (which backs up the mapping of files and directories' mapping).
HDFS clients create a new document by informing its route to the NameNode. For every block of the document, the NameNode feedbacks a list of DataNodes to handle its replicas. Then, the clients pipeline data to the selected DataNodes,
where, in the end, the latter confirms the production of block replicas to the NameNode.
HDFS requires the NameNode with limited size to store all the NameSpace and block positions, so that the quantity of addressable blocks is limited, leading to the scalability problem of the NameNode.
The framework of HDFS is illustrated in Fig. \ref{HDFS}.

Hadoop needs to write the program state to disk per iteration, therefore its performance on a number of ML programs has been surpassed by in-memory alternative platforms. One alternative is Spark \cite{zaharia2010spark}, which keeps the ML program state in memory and leads to significant performance gains in comparison with Hadoop, while keeping the interface of MapReduce simple and easily usable.
In \cite{Gu2018Parallelizing}, a DML optimization framework is proposed on top of Apache Spark by applying the parameter server framework. Distributed synchronous Elastic Averaging
SGD (EASGD) and some other classic SGD-based methods are designed and evaluated on the platform. To obtain the optimal mini-batch size, Gu \emph{et al.} \cite{Gu2018Parallelizing} also empirically analyze the famous linear scaling rule, which intends to calibrate the learning rate linearly in accordance with the mini-batch size.
It is shown that the linear scaling rule follows with a limited mini-batch size.
Spark ML implementations are usually slower than
specialized ones, partially because Spark
does not schedule computations and communications flexibly in a fine-grained interval. It has been shown that proper scheduling is very important to the fast and correct implementation of ML programs.

A Simple Scalable Streaming System (S4) is developed as a general-purpose ML platform for executing continuous unbounded data streams in \cite{neumeyer2010s4}. It offers a scalable distributed stream processing engine, while MapReduce operates only batch computing. In S4, events with keyed attributes are transferred to Processing Elements (PEs). PEs perform the following operations upon the events: (i) emitting the events that may be consumed by the other PEs, and (ii) publishing the processing results.
The architecture includes the Actors model \cite{agha1985actors}, and provides the semantics of encapsulation and location transparency. Hence, the applications can run simultaneously. A simple programming interface is provided to application and software developers. Neumeyer \emph{et al.} \cite{neumeyer2010s4} outline the S4 architecture and describe a range of possible applications, including real-world deployments.
The S4 design is shown to be flexible and suitable to run in large clusters built with commodity hardware.
The proposed S4 system in \cite{neumeyer2010s4} exploits static routing and load balancing, and lacks robust live PE migration.

While external storage buffering big data can prolong the training period during deep learning,
Lee \emph{et al.} \cite{Lee2017BAIPAS} propose a so-called ``Big Data and AI-based Predication and Analysis System (BAIPAS)'', to speed up training using big data.
By taking into consideration the data size, the availability of the server's memory and CPU/GPU, the BAIPAS system analyzes the states of the training data and the worker servers, and then evenly distributes the training data among all the worker servers. Data shuffling is also performed to move data between servers during training, so that each server (with only a subset of the entire training data at the beginning) can learn from the entire data set to avoid model over-fitting.

Theoretical insights are provided on how coded techniques can achieve significant gains over the uncoded in \cite{Lee2018Speeding},
in terms of robustness against system noise such as stragglers and communication bottlenecks~\cite{dean2013tail}.
An encrypted calculation framework (using erasure codes) is proposed to alleviate the effect of stragglers
by introducing redundancy into submissions of a decentralized approach,
and obtaining the computing outcome which is decrypted from a subset of the submission outcomes,
dropping unfinished submissions.
The coded computation speeds up distributed matrix multiplication by a coefficient of $\log n$, with $n$ denoting the number of workers.
Coded Shuffling is further proposed to significantly decrease the communication overheads of data-shuffling,
which is demanded for the realization of high mathematical efficiency in DML approaches.
The key idea of coded shuffling is to multicast a coded common message to workers, instead of unicasting multiple separate messages.

\begin{figure*}
\centering
\includegraphics[width=0.7\textwidth]{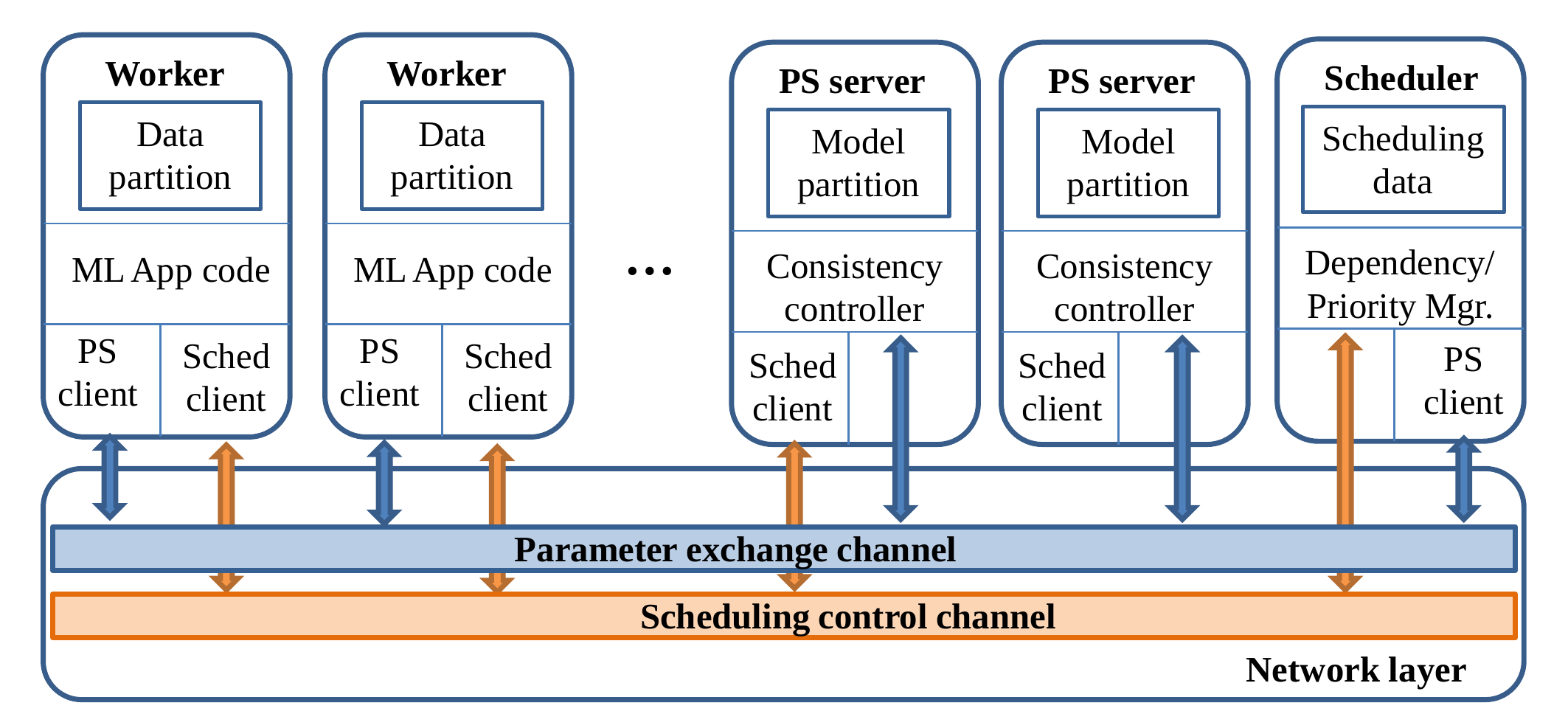}
\caption{An illustration of the Petuum system including a scheduler, multiple workers and parameter servers \cite{xing2015petuum}.
The central system follows the parameter server architecture, allowing developers to access the global ML model from every node through a simple interface to the distributed and shared memory of the system.
The system also quantifies the inconsistency resulting from asynchronous operations to guarantee the convergence and stability of the resulting ML model. }
\label{Petuum}
\end{figure*}

\subsection{Parameter Server-Based Software Platform}
A general-purpose DML platform, Petuum, is designed
based on an ML-centric optimization-theoretic principle in \cite{xing2015petuum}.
Petuum aims to operate iterative updates of advanced ML algorithms to quickly
converge to an optimum of the objective function. The goal is achieved by exploiting three statistical properties enabling efficiency in large-scale distributed
ML: Error tolerance, dynamic structural dependency, and non-uniform convergence. Correspondingly, Petuum introduces three new system objectives: (i) Petuum can synchronize the parameter
states and avoid data staleness, thus achieving right results at a much lower cost than traditional bulk synchronization per iteration; (ii) Petuum provides dynamic schedules by considering the time-varying structural dependencies among system parameters, such that loss of parallelization and synchronization can be minimized; and (iii) Petuum can give priority to any non-convergent parameter settings to achieve quicker convergence since ML parameters have different convergence rate.

Petuum offers APIs to central systems to make easy access of data- and model-parallel computing, see Fig. \ref{Petuum}. The central system follows the PS architecture, allowing developers to access the global ML model from every node through a simple interface to the distributed and shared memory of the system. This interface resembles single-machine programming. The system also quantifies the inconsistency resulting from asynchronous operations to guarantee the convergence and stability of the resulting ML model. Hence, explicit network synchronization is omitted. Petuum also contains a scheduler, which allows its users to determine their individual principles of the consistency of applications \cite{xing2015petuum}.
The platform can still be improved in terms of fault recovery from partial program state, as well as its feasibility to adjust resource allocation in real-time.

A shared memory-based DNN framework is proposed in \cite{Lim2017accelerating} to accelerate the process of reading and updating parameters in a PS architecture. In particular, a remote shared memory which can be accessed across multiple workers, is used to maintain global shared parameters of parallel workers. Simulation shows that, compared with TensorFlow, the training time of the proposed framework for image recognition is saved by $10\sim50$\% when training CNN and Multi-Layer perception (MLP) models.

\subsection{Graph-Based Platform}
Graph-based platforms, such as Pregel \cite{malewicz2010pregel} and GraphLab \cite{low2012distributed}, effectively divide graph-based models with embedded scheduler and consistency controller. To address the distributed processing of large scale graphs, Malewicz \emph{et al.} \cite{malewicz2010pregel} built Pregel, a scalable platform with an API to support many graph algorithms. However, little to no analysis has been carried out to confirm that the asynchronous consistency models and scheduling of graph-based platforms can always correctly execute ML tasks.
The limitations of popular software platforms are summarized in Table \ref{tab.software}.


\section{Applications of DML to Wireless Networks}\label{sec.use}

\begin{table*}[t]
\renewcommand{\arraystretch}{1.5}
\caption{Applications of Distributed Machines Learning Techniques in Wireless Networks}
\centering
\begin{tabular}{|c|c|c|c|c|}
\hline

\diagbox{Techniques}{Applications}&Power control & Spectrum management  & QoS provision & Resource allocation \\ \hline
Federated learning  &\cite{tung20, yang18, cao19, tao20, feng18}   &\cite{kaibin19}
&\cite{ha19, haba19, yang20, moham20} 
&\cite{mahan17, chen18, kone15, sama18, chen20, xujie20} \\ \hline
Federated reinforcement learning  &--   &\cite{moroz14, yan20}      &--  &\cite{marco15} \\ \hline
Federated deep learning         &--   &--   &--  &\cite{yu19,yuris19} \\ \hline
Partitioned learning  &--   &--   &--  &\cite{wen20} \\ \hline

\end{tabular}
\end{table*}

In large-scale wireless networks with massive datasets, optimization methods, such as convex optimization, dynamic programming and (sub-)gradient descent-based approaches may yield a high computational complexity and a slow convergence to (sub-)optimal solutions.
Although data-driven ML techniques can solve the problems faster,  they call for more powerful computation capability and storage
than the mobile devices can generally have~\cite{gunduz19}.
One solution to overcome this limitation is to establish a cloud or edge unit to collect all the data from the wireless devices
and train an ML algorithm in a centralized fashion.
Yet this approach is further constrained by the bandwidth, channel condition, latency, energy consumptions,
as well as privacy concerns.
Therefore, DML frameworks, such as FL and partitioned learning,
have been proposed that allow wireless devices to acquire a global model with limited data exchange
or based on partial models and datasets, such as \cite{tung20, yang18, wen20}, and \cite{cao19}.

\subsection{Power Control}
A new scheme is proposed in \cite{tung20} for cell-free massive multiple-input multiple-output (CFmMIMO) systems
to enable FL frameworks.
The approach permits each of the iterations of FL to take place in a long coherence period to make sure that
the FL operation is stable.
The joint optimization of local precision, transmit power, throughput, and users' working frequency is formulated as a mixed-timescale stochastic non-convex problem of power control.
This captures the sophisticated interoperations among the training duration,
and the communication and mathematical operation of weight refreshment of an FL process.
By using an online successive convex approximation technique, the power control problem is solved iteratively with ensured convergence to within the neighborhood of its stationary solution.
Simulations validate that the hybrid optimization diminishes the training time by up to $55\%$,
in contrast to several benchmark approaches, namely,
(a) equal downloading (DnL) power allocation to all user equipments with the maximum uploading (UL) transmit power and fixed local accuracy;
(b) equal power allocation with the maximum UL transmit power and local accuracy;
and (c) equal power allocation with a fixed local accuracy and the DnL and UL transmit powers.
It is shown that CFmMIMO consumes the shortest training duration of FL, in contrast to
cell-free time-division multiple-access (TDMA) massive MIMO and co-located massive MIMO.
Yet, the results are obtained by assuming time-invariant channel state information,
which may limit the application of the algorithm.

\begin{figure*}
\centering
\includegraphics[width=0.6\textwidth]{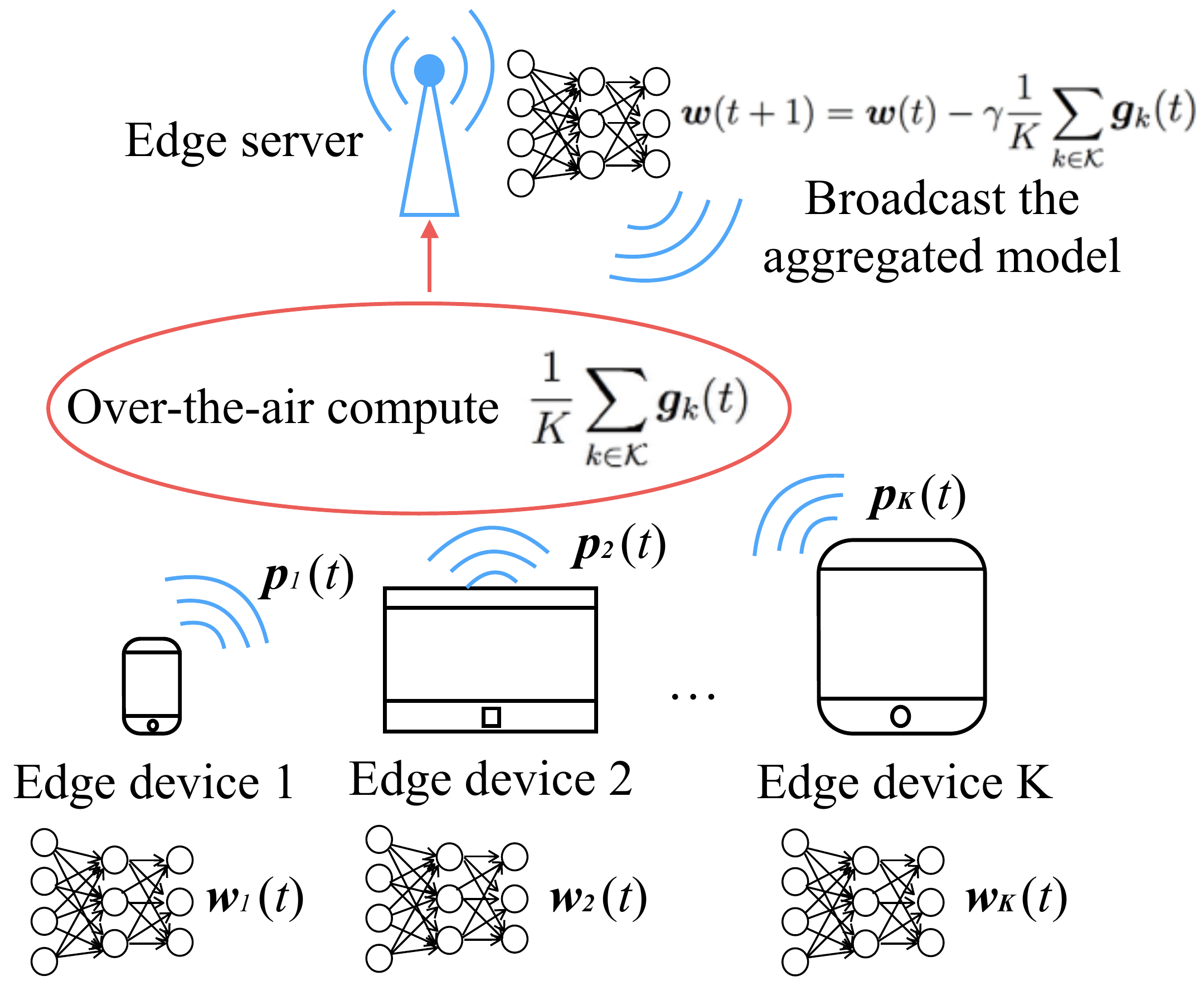}
\caption{An illustration of the over-the-air FL \cite{tao20}, where $k \in \{1, \cdots, K\}$ is the index of edge devices,
$\boldsymbol w$ is the model parameter, $p$ is the transmit power, $\gamma$ is the learning rate, and $\boldsymbol g_k(t)$ is the gradient of the loss function computed at device $k$ at time block $t$.}
\label{tao20air}
\end{figure*}


Power management at edge devices (or agents) is critical to realize a dependable and well-behaved AirComp over fading channels.
Existing works focus on conventional data collection, which typically assumes that the data gathered locally by various agents
are i.i.d.
The data are normalized to be a zero-mean Gaussian process with a unit variance~\cite{yang18, cao19}.
Yet, such an assumption cannot be applied to gradient collection in ML because
the gradient distribution may not be identical between iterations,
and the statistical characteristics of each entry of the gradient vector may differ substantially even in the same iteration.
Zhang and Tao \cite{tao20} optimize power management for over-the-air FL by considering the gradient statistics,
as shown in Fig. \ref{tao20air}.
The target is to achieve the minimum model gathering inaccuracy quantified based on mean square error (MSE).
This is realized by collaboratively acquiring the best transmit power of each agent and a de-noising coefficient at the edge server.
A closed-form optimal solution is devised, when the first- and second-order statistics of the gradient are  {\em a priori} available. 
The optimal solution relies on a multivariate parameter of the variations of the gradients.
The statistics of the gradient are estimated based on the historically collected gradients.
The statistics are later used to successively modify the transmit powers of the devices over each learning iteration.
Experimental results validate that the power management outperforms a straightforward full-power communication approach and a threshold-based power management approach in terms of model precision and convergence speed.

The authors of~\cite{feng18} construct a cooperative FL communication platform with relays to support model updating, transmit and trading.
Within the platform, mobile devices produce model renewals pertaining to their locally acquired or generated sample data.
These model renewals can be transferred to a model master via a collaborative relay network.
The model master benefits from the training task offered by the mobile devices which, in return, charge the model master certain fees.
Given the strong interferences of wireless communication among the devices using a single relay node,
sensible devices need to choose their (different) relay nodes and configure their transmit powers.
A Stackelberg game model is constructed to describe the reciprocal actions among the devices
and the interactions between the devices and the model master.
The Stackelberg equilibrium is achieved by employing the exterior point method.

\subsection{Spectrum Management}

Edge ML consists of training techniques developed at the edge of the system to take advantage of enormous distributed datasets
and mathematical operation resources.
The architecture of federated edge learning (FEEL) has attracted significant attention for its capability of preserving data privacy.
FEEL schedules the overall model learning at each individual server,
schedules the local model learning at each of the edge agents,
as well as the model synchronization between the server and edge devices through wireless links.

Bandwidth allocation and coordination are optimized in \cite{kaibin19} to achieve the least energy usage of mobile devices.
The optimized strategies are adaptable to the channel states and computation capabilities of the devices
to lessen the total energy and warrant the training effect.
Compared to conventional throughput-maximization schemes \cite{zhou18mec},
the optimal strategies assign wider spectra to individuals with either poorer channels or more deficient computing powers.
Both poor channel conditions and deficient computation
are major drawbacks of simultaneous model renewals in FEEL.
A weight function rewards the agents
with stronger channels and higher mathematical operation capabilities.
The experiments show that significant energy saving can be achieved.
One limitation of the system, however, is its difficulty to be generalized to the application of other model transmission cases, such as asynchronous models.

\begin{figure*}
\centering
\includegraphics[width=0.6\textwidth]{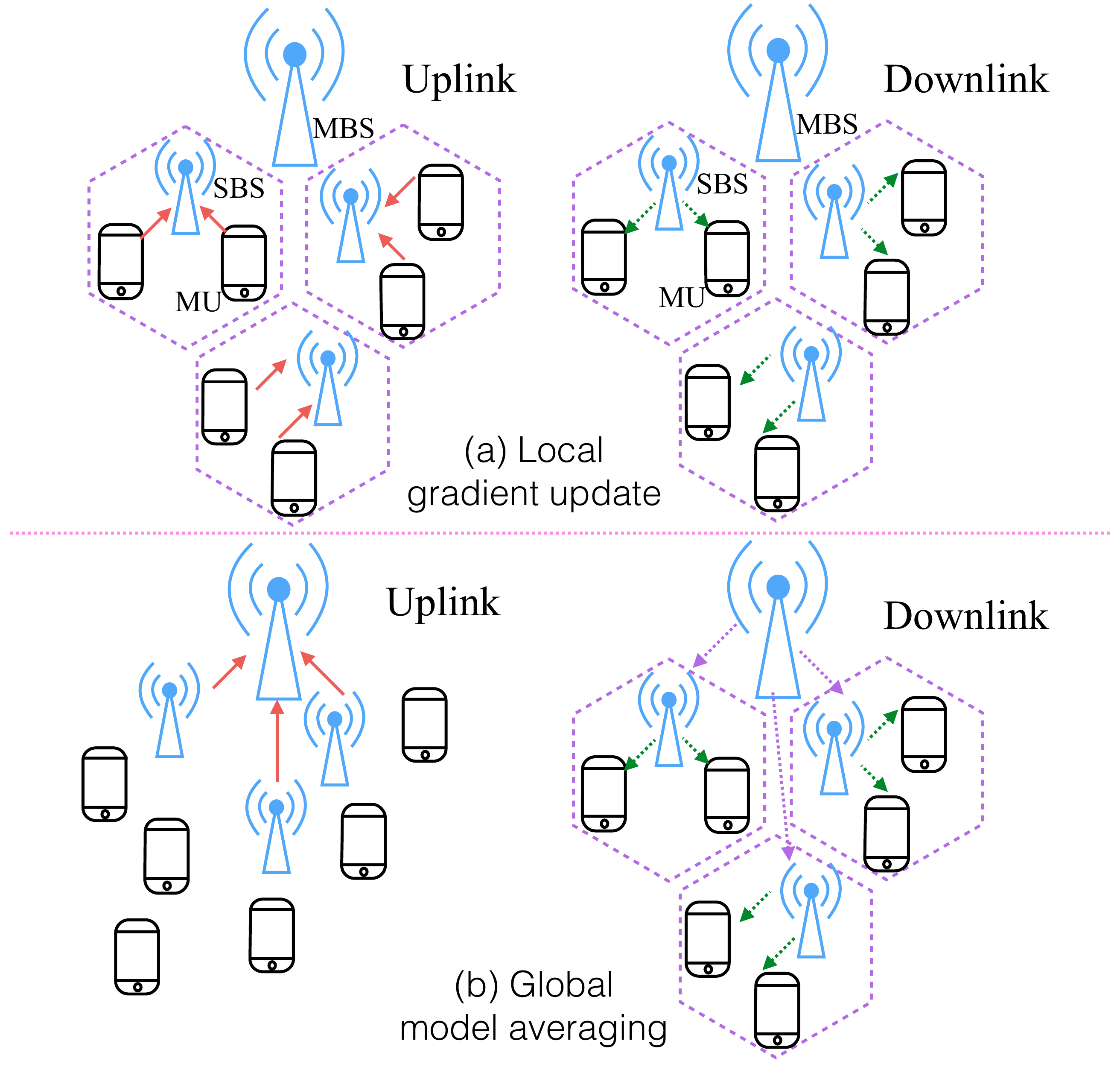}
\caption{A hierarchical FL system model \cite{abad19},
where small BSs (SBSs), such as micro- or pico-BSs, carry out FL amongst many mobile users inside the radio coverage of the SBSs.
The SBSs repeatedly communicate the resultant model renewals to a macro-BS in order to reach network-wide agreement.}
\label{abad19model}
\end{figure*}

FEEL is leveraged in a heterogeneous cellular network in~\cite{abad19},
in which micro-BSs or pico-BSs carry out FL amongst many mobile users inside the radio coverage of the BSs.
The BSs repeatedly communicate the resultant model renewals to a macro-BS in order to reach a network-wide agreement.
Gradient sparsification is employed in conjunction with periodically averaging to make the hierarchical FL efficient in terms of communication.
A sparse gradient vector allows the transmission of only a portion of the parameters in each round,
thus reducing the communication latency.
Image processing is performed by using CIFAR-10 dataset that includes $60,000$ $32\times32$ color pictures in $10$ categories,
with $6,000$ pictures per category.
There are a total of $50,000$ training pictures and $10,000$ test pictures~\cite{cstoron}.
 By utilizing CIFAR-10 dataset, it is shown that, this hierarchical training approach
is able to prominently decrease transmission delay without sacrificing the model precision.

The idea of ``Win-or-Learn-Fast (WoLF) changeable learning rate'' is developed in \cite{moroz14} to be part of Q-learning to manage spectrum allocation and usage in a decentralized fashion.
In this concept, a participating device should accelerate the learning process when it is losing and slow it down when winning.
The WoLF learning rate is evaluated by separating the learning rate $\alpha$ into $\alpha_{\text{win}}$ and
$\alpha_{\text{lose}}$, when the agent succeeds and fails in file transmissions, respectively.
If $\alpha_{\text{win}} < \alpha_{\text{lose}}$, the WoLF rule is satisfied, because the device learns more slowly on successful attempts
and more quickly on failed attempts.
The significance of selecting the right learning rate is demonstrated numerically by testing a large-scale stadium event network,
in which a wireless network is established in the stadium to temporarily boost mobile throughput for users attending the event
(such as a football game).
The simulations demonstrate that by adopting the WoLF learning rate (e.g., $\alpha_{\text{win}} =0.01$ and $\alpha_{\text{lose}}=0.05$)
a substantial enhancement is obtained in terms of file transmission failure probability and stoppage probability,
as compared to an unchanged learning rate of $0.1$.

\subsection{QoS Provisioning}

In large-scale wireless networks, the reduced coupling among the training participants can accelerate the convergence of the training process.
Therefore, decentralized multi-agent Q-learning has been applied to spectrum sensing and radio resource management (RRM) problem
when the BSs (or agents) do not need to acquire the explicit strategies of each other~\cite{sun19, marco15}.
The intelligent and uncoordinated BSs can make decisions based on different local input data (such as users' distribution and traffic load)
and partial observations of the environmental state.
As an example, consider the RRM problem  at a renewable energy source (RES)-powered BS~\cite{marco15}.
The policy-choosing procedure of a BS is described as a Markov Decision Process (MDP).
The state vector of the MDP is
$\boldsymbol s := \{s_i\}_{i=1}^I$, where $s_i$ is the state of BS $i$ (i.e., its battery level).
Based on $s_i$, each BS selects \emph{independently} an operation $a_i$ from the operation set to decide its ON or OFF operation state.
As a result, the environment returns an \emph{agent-dependent} reward $r_i$ to capture the system throughput and battery level,
facilitating a local update of the Q value.

FL algorithms are utilized to achieve the shortest computation and communication latency in \cite{ha19},
and for traffic prediction to achieve the largest throughput of users in \cite{haba19}.
Unlike many centralized learning techniques typically running in data centers,
FL is suitable for wireless edge networks,
where the communication channels are restrained and lossy.
Given a relatively limited wireless bandwidth, only some users can be coordinated for model (or weight) renewals at each iteration.
For instance, only those with their signal-to-interference-plus-noise ratios (SINRs) satisfying some specific predefined requirements (in other words, with comparatively reliable channels) are selected to contribute to the model updates~\cite{yang19FL}.
Wireless transmissions are also influenced by interference.
In \cite{yang20}, an algebraic model is established which describes the effect of FL in wireless systems.
Mathematical expressions are developed to evaluate the convergence speed of FL,
analyzing the effects of both coordination and inter-cell interference.
A comparison study is carried out with three different coordinating strategies, namely, random scheduling, round-robin,
and proportional fair, with regards to FL convergence speed.
FL is validated to be more efficient running with proportional fair than with random scheduling and round-robin,
if the system operates under a high SINR target.
Round-robin is favorable if the SINR target is low.
Additionally, the FL convergence speed drops dramatically as the SINR target rises,
which confirms the significance of reducing the size of the renewed parameters.
Moreover, the mathematical analysis unveils a trade-off between the numbers of coordinated users and the subchannel bandwidth,
given a usable amount of spectrum.

Amiri and G$\ddot{\text u}$nd$\ddot{\text u}$z \cite{moham20} study FL at wireless network edges,
in which power-constrained wireless agents each owning a private dataset establish a mutual model under the coordination of a distant parameter server (PS).
A bandwidth-constrained fading multiple access channel (MAC) protocol is developed between the participating agents and the PS,
to utilize decentralized stochastic gradient descent (DSGD) in a wireless fashion.
A digital DSGD (D-DSGD) approach is first proposed, where one agent is chosen for each iteration
by taking advantage of the devices' channel states.
The coordinated agent discretizes (or quantizes) its gradient estimation depending on its channel state.
The agent returns to the PS the quantized bits.
By making use of the additive attribute of wireless MAC, a novel analog transmission approach,
namely, compressed analog DSGD (CA-DSGD), is created in which the agents first extract sparse estimations of their gradients while gathering errors from past iterations.
Then, the devices produce a projection of the obtained sparse vector into a vector space with lower dimensions.
The authors of \cite{moham20} also develop a power assignment policy to achieve the consistency and alignment between the obtained gradient vectors from the perspective of the PS.
Simulation results demonstrate that CA-DSGD converges much more quickly than D-DSGD with significantly better accuracy.

\subsection{Resource Allocation}

\begin{table*}[t]
\renewcommand{\arraystretch}{1.5}
\caption{Distributed Machines Learning Techniques for Wireless Resource Allocation}
\centering
\begin{tabular}{|c|c|c|c|c|}
\hline

\diagbox{Techniques}{Applications}&User association & \makecell{Power allocation \\ and latency} 
& \makecell{Bandwidth allocation \\ and user selection} & Edge cloud computing \\ \hline
Federated learning  &\cite{chen18}   &\cite{sama18} &\cite{chen20, amiri20} 
&\cite{xujie20} \\ \hline
Federated reinforcement learning  &--   &--      &--  &\cite{marco15} \\ \hline
Federated deep learning         &--   &--   &--  &\cite{yuris19} \\ \hline
Partitioned learning  &--   &--   &--  &\cite{wen20} \\ \hline

\end{tabular}
\end{table*}

\begin{figure*}
\centering
\includegraphics[width=0.8\textwidth]{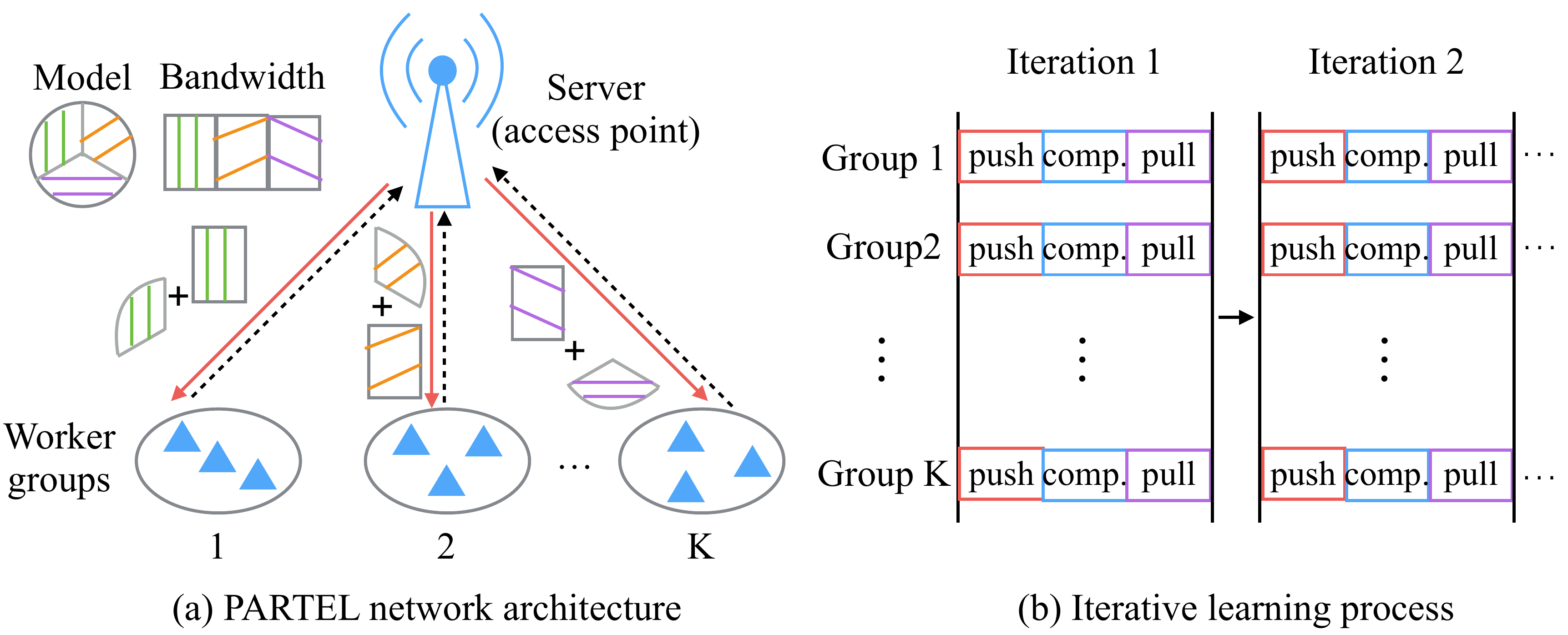}
\caption{System model and operations of the PARTEL framework \cite{wen20}.
(a): The model is adaptively separated into different blocks of parameters that are downloaded to agent clusters
to refresh sample data subsets with allocated bandwidth.
(b): At each iteration, all the parametric blocks are updated.}
\label{wen20partel}
\end{figure*}

\subsubsection{User association}
An echo state network (ESN)-based FL approach is investigated in \cite{chen18}, which estimates the positions
and poses of wireless virtual reality users.
This algorithm utilizes FL to allow several BSs to learn and update their deep ESNs locally.
A learning model is created cooperatively to estimate the locations and orientations of all users,
after being trained with practically captured datasets.

\subsubsection{Power allocation and latency}
Joint power and resource allocation is studied in \cite{sama18} for ultra-reliable low-latency communication (URLLC) with a specific application to vehicular networks.
To minimize queueing delay, FL is exploited to decentralize model training and operations with a focus on accurately modeling the queues based on their tail distribution.
However, neither of these studies \cite{chen18, sama18} considers the lossy nature and limited bandwidth of wireless channels,
both of which can compromise the quality, performance, and convergence of FL.

\subsubsection{Bandwidth allocation and user selection}
FL algorithms are trained over a practical wireless channel in \cite{chen20},
where the shortage of bandwidth during the transmission of the FL model is investigated.
Since all the learning parameters are delivered via wireless links, the effect of the learning is influenced by the packet errors and the limited and time-varying availability of wireless channel capacity.
The BS can only select a subgroup of users to orchestrate the FL at any moment.
Learning radio resource assignment and user pairing are jointly done to minimize the FL loss function.
A closed-form expression is obtained for the convergence speed of the FL
and to analytically evaluate the influence of the model parameters.
With an estimated convergence speed, the best transmit power of each user is observed to depend on
user selection and radio channel assignment.
The best user selection and radio channel assignment can be achieved by running a convex search.
Numerical simulations  based on Matlab demonstrate that the proposed hybrid framework of FL and communication
can decrease the loss function of the FL by at most $10\%$ and $16\%$, respectively,
compared to i) the best user selection with stochastic resource assignment,
and ii) the standard FL with stochastic user choosing and resource assignment.
The results demonstrate that a balanced consideration of learning configuration and resource allocation is important for the accuracy and efficiency of FL.

The potential use of FL at the wireless edge is investigated in~\cite{amiri20},
where power-constrained agents with local datasets jointly learn a model facilitated by a distant PS.
The learning task considered is to minimize an empirical loss function.
The agents are linked to a PS via a shared, bandwidth-constrained wireless link.
At each iterative cycle of FL, a subgroup of agents are coordinated and transfer their local model renewals to the PS by using orthogonally assigned frequency channels.
Each engaging agent needs to reduce the size of its model renewal to meet its channel capacity.
The proposed coordinating strategy captures both the link states and the prominence of any local model renewals.
It yields a more significant and more prolonged effect in comparison with strategies capturing solely either of the two measurements.
It is revealed in~\cite{amiri20} that when the data is i.i.d. across agents,
choosing a single agent for communication at each round yields the ideal effect.
In the presence of non-i.i.d. data, coordinating several agents at each round enhances the learning effect.
Therefore, the number of coordinated agents should rise for less diversified and impartial data distribution.

\subsubsection{Edge cloud computing}
Mo and Xu \cite{xujie20} design a FEEL system, where an edge server schedules multiple wireless edge agents
to learn an ML model collectively using their locally collected data samples.
During the decentralized learning, communication and computation are optimized holistically to improve the system energy efficiency.
Two wireless transmission protocols are developed for the agents to transfer the model parameters to the server, i.e.,
by using non-orthogonal multiple access (NOMA) or TDMA.
Given a training accuracy requirement, the sum energy usage of all edge agents is minimized over a limited learning period
by convexifying and optimizing the transmit power and speeds at the agents for parameter transmission, as well as their CPU configuration for local updating.
Numerical results demonstrate that the collaborative communication and computation optimization
enhances the energy efficiency of FEEL
by balancing the tradeoff of energy between communication and computation, as compared to its alternatives
only optimizing either communication or computation.
However, the joint optimization involves approximation for convexification.

The framework of FEEL does not partition models, and demands each edge agent to refresh an entire model.
Because the agents can often be resource-limited, FEEL is typically applicable to small or moderate training missions.
By contrast, an important goal (or objective) of partitioned edge learning (PARTEL) is typically to learn large-scale models employing numerous agents as workers via model partitioning.
In other words, a global model is split into many much smaller models,
each of which accounts for a portion of the original learning model~\cite{limu13}.
Hence, the design of efficient PARTEL demands the decoupling of load assignment from radio resource assignment,
inviting distinct challenges not encountered by FEEL.

Wen \emph{et al.} \cite{wen20} consider the PARTEL framework to iteratively learn a large-scale model with numerous resource-limited wireless agents (referred to as ``workers'').
In each communication round, the model is adaptively separated into different blocks of parameters
that are downloaded to agent clusters to refresh sample data subsets.
At the next step, any local renewals are sent to a designated server at which the renewals are managed to amend an overall model.
To minimize the total learning-and-communication latency and save resources, the parameter (i.e., calculation load) assignment and radio bandwidth assignment (for downloading and uploading) are jointly pursued.
Two operating methods are employed.
\begin{itemize}
\item[$\bullet$]
The first one is a practical successive method, namely, partially integrated parameter-and-bandwidth allocation (PABA) method.
The approach consists of two schemes, referred to as bandwidth-aware or parameter-aware allocation of spectrum.
The bandwidth-aware allocation assigns the least load to the slowest (in terms of computing) of agent clusters,
each learning the same block of parameters.
The parameter-aware bandwidth allocation assigns the widest bandwidth to the particular worker which has been the latency bottleneck.

\item[$\bullet$]
The second approach is to collaboratively optimize PABA.
Despite being a non-convex problem, a fast and optimal approach is developed by using nesting bisection search,
and convexified to be efficiently solved with a convex program.
The decentralized training under the PARTEL framework is optimal since it achieves the same training effect (in terms of convergence rate) as the centralized training within the same number of iterations.
Experimental results are obtained by learning a news-filtering model by the News20 data aggregated in~\cite{news20},
and show that integrating PABA can significantly improve PARTEL in terms of latency (e.g., by $46\%$)
and accuracy (e.g., by $4\%$).
\end{itemize}

Enabling ML at wireless edge (for instance, edge cloud) near mobile terminals and devices is crucial for many next-generation mobile and IoT applications.
It is also challenging because the lower layers of an edge cloud significantly differ from current ML configurations.
In a geo-decentralized operating system, streaming data needs to be decomposed at a low cost,
meanwhile reserving the representativeness of the decomposed data for different workers.
This can be vital to generate an impartial training model whose parameters are updated simultaneously at a practical interval.
Lyu \emph{et al.} \cite{lyu19} develop a real-time method to optimally decompose continuous data streams under dynamic system states over time.
A new metric quantifies the fairness of data decomposition (to maintain the characteristics of the decomposed data),
and is used to constrain the design of data permission, decomposition, examination and analysis.
SGD is utilized to obtain the best policies in real-time and achieve asymptotically the largest gain of data decomposition.
Numerical tests validate that the developed scheme outperforms the classic scheme on the aspects of data rate and cost-effectiveness,
whereas only $24\%$ of the system connections have to be evaluated without the cost of the asymptotic optimality
in their considered simulation setting.

The authors of \cite{yuris19} propose two new proactive cooperative caching schemes by utilizing DL to estimate the content requirement of users in a wireless caching system.
In one of the schemes, a content server (CS) is responsible for collecting data from every mobile edge node (MEN).
The CS also performs DL to predict the content required for the system.
The centralized scheme is susceptible to the leak of individual (personal) data due to the MENs' share their personal information with the CS.
In the second scheme, a decentralized DL framework is proposed.
The distributed DL permits the MENs to cooperate and exchange data for diminishing the error of the content requirement estimation
without releasing the individual data of the users.
Numerical results validate that the developed schemes can refine the precision by decreasing the root mean squared error (RMSE) at most $33.7\%$.
These schemes also shorten the service latency by $47.4\%$, in contrast to several other ML approaches.


\begin{table*}[t]
\renewcommand{\arraystretch}{1.5}
\caption{Techniques and Their Implementations for Data Privacy and Security in Distributed Learning Systems}
\centering
\begin{tabular}{|c|c|c|c|}
\hline
Privacy at agent &Perturbation & Dummy  & Encoding   \\ \hline
References  &\cite{dwork06, geyer17}   &\cite{kido05, dummy16} &\cite{Zhang2019Privacy} \\ \hline
Privacy at server  &Aggregation   &Secure multi-party computation      &--   \\ \hline
References         & \cite{gu19, Jia2019Efficient}   &\cite{phong18}   &--  \\ \hline
Implementations  &Prevention of data poisoning   &Perturbation plus encryption   &Blockchain-based solutions   \\ \hline
References     & \cite{zhang20, Zhang2018A}  & \cite{hao20}   & \cite{kim20, Yang2018byrdie, Yang2019byrdie}   \\ \hline
\end{tabular}
\end{table*}

\begin{figure*}
\centering
\includegraphics[width=0.8\textwidth]{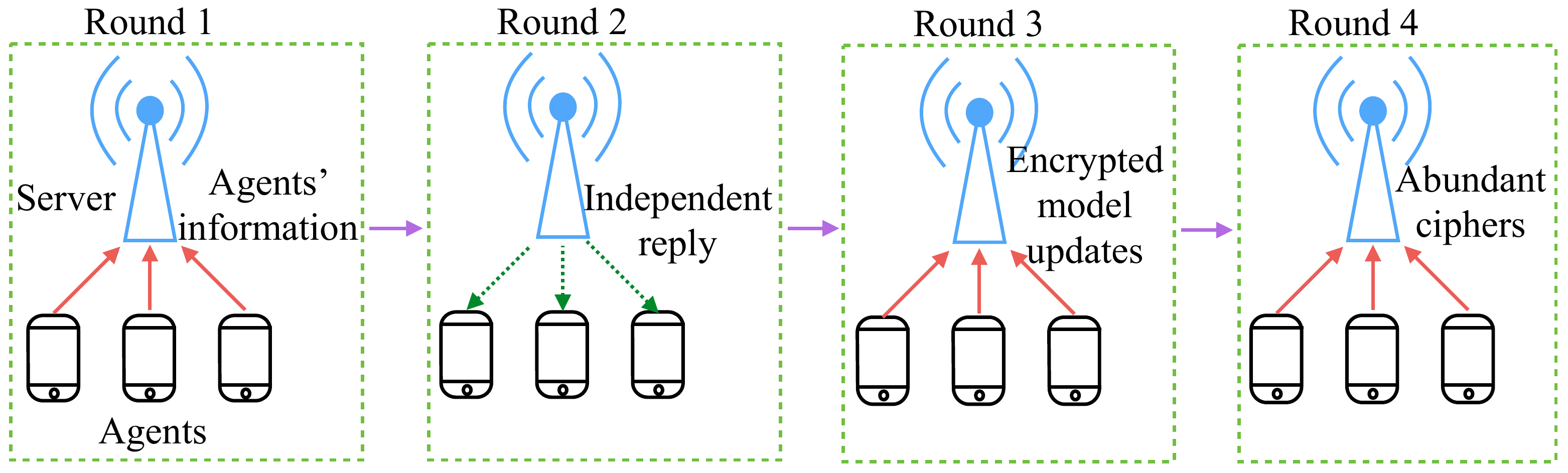}
\caption{An illustration of the four-round encryption of secure multi-party computation.
The first round starts with the server gathering information from all agents, followed by using the collection of agent information
to obtain an independent reply and send back to each agent in the second round.
The devices upload encrypted and masked model renewals to the server in the third round.
The fourth round is the finalization stage where the devices release abundant ciphers to enable the server
to decrypt the collected model renewals.}
\label{smcfig}
\end{figure*}

\section{Privacy and Security}

DML algorithms are expected to play an important part in examining and analyzing numerous datasets in large-scale systems~\cite{dai18}.
With ``data-hungry'' learning algorithms and the exploding information,
it is crucial to guarantee data privacy and security for the distributed learning system against a third party,
or even against other devices within the system.
Yet, the increasing dependence on ML techniques renders it intrinsically susceptible to cyberattacks,
such as malicious servers, eavesdroppers, and data poisoning~\cite{ma20, miller20}.
In this section, we review popular techniques and effective countermeasures that can be utilized by the agents and the servers
to preserve data privacy and security, and their implementations targeting different attacks.

\subsection{Privacy of Agent}
In FL, the agents forward their training outcomes, i.e., the parameter values and weights, to a server.
In practice, the agents might not have the trust in the server because the server could potentially dictate and manipulate the training process, and extract private information of the agent based on values made available to the server.
To address this issue, agents can leverage some of the following privacy-preserving techniques:

\subsubsection{Perturbation}
The concept of perturbation is to append artificial noises to the parameters uploaded by the clients.
Differential privacy \cite{dwork06} is often studied to veil certain sensitive features until no third-party can
differentiate individuals, the data is impossible to be restored, and the user privacy is preserved.
Geyer \emph{et al.} \cite{geyer17} present a differential privacy scheme to FL to further safeguard agent-side information.
The authors intend to conceal agents' contributions during training, and strike a balance between the privacy loss and learning performance.
Experimental results demonstrate that given enough engaging agents,
the proposed approach can preserve client-side differential privacy at a little cost in model performance.

\subsubsection{Dummy}
A reliable method of preserving data privacy at the client-side is to send dummy model parameters along with the true ones
to the server to hide the clients' individual contributions during training.
In particular, Kido \emph{et al.} \cite{kido05} propose a dummy scheme where a user of a location-aware service sends multiple counterfeit location information (dummies) to the service supplier, together with true location information.
Since the service supplier cannot extract the true location information, the user's position privacy is guarded.
Augmented dummy data bears redundancy, requires extra bandwidth and buffer, and consumes more energy.
Diyanat \emph{et al.} \cite{dummy16} minimally augment dummy data to preserve the (original) data privacy of a client without changing the statistical behavior of the original data, such as distribution.
They minimize the weighted sum of augmented dummy communication cost and privacy degree.
It is guaranteed that the (malicious or curious) server's estimation (or speculation) of the true data
has a probability of error higher than a certain threshold.

\subsubsection{Encoding}
In \cite{Zhang2019Privacy}, a DML structure is developed with an encoder installed at every data owner (or in other words, worker)
for training data and protecting the privacy of the data occupiers.
The objective of the encoder is to encrypt the collective characteristics of data,
and work as the relay to transmit the characteristics to a centrally-located server, instead of the raw data of the users.
The encoder obtains high-level functions from admitted data, and forwards the characteristics to the central server.
This design prevents sharing raw data in the system.
In the example given in \cite{Zhang2019Privacy}, the encoder extracts the image characteristics, and does not expose the image itself.

\subsection{Privacy of Server}
Upon gathering all parameters from many agents, the server performs a weighted average of the parameters.
The weights typically depend on the size of the data which the clients train to update their parameters.
Yet, when the server releases the collected parameters for the model synchronization between the agents,
the model parameters can potentially be leaked to passive eavesdroppers in the network~\cite{shen19jiot}.
To this end, protections of the server-side are of importance.
The following methods have been utilized to protect privacy at the server side.

\subsubsection{Aggregation}
Essentially, the aggregation allows the server to collect data or parameters originated from various agents.
The gathering can be accomplished by a PS.
Since the number of updates received from the clients can be very large, it is inefficient and even impossible
to apply all updates to the global model one by one.
The main purpose of the data aggregation process is to extract and make full use of the data contained in the updates.
This process takes all received updates as inputs. Its output can be either a final update which can be directly applied to the global model,
or a new model which replaces the old one~\cite{gu19}.
In some other cases, the servers are free to choose the agents with high-quality parameters
or easily-satisfied demands~\cite{gu19}.
For FL, how to devise a suitable collection mechanism is an open problem.

A hierarchical distributed system is developed in \cite{Jia2019Efficient} with different solutions for different data partitioning scenarios.
In a hierarchical distributed mode, all entities can be divided into different layers.
The lower layer corresponds to the actual data owners.
The upper layer consists of nodes collecting data.
With an asynchronous learning strategy, the hierarchical architecture has a flexible choice of model parameters, and also a shorter computation time before convergence.
The design cannot provide an optimal solution for data exposure.
Also, it is unable to offer the optimal choice of the step sizes for convergence speed control parameters,
which could potentially lead to worse classification performances.

\subsubsection{Secure Multi-Party Computation (SMC)}
In \cite{rosulek17}, the individual devices (or agents) try to prevent the servers from inspecting their updates by only reporting the sum after a sufficiently large number of updates.
The SMC adopts encryption techniques to prevent the inspections in the following four rounds during a communication stage of FL,
as illustrated in Fig. \ref{smcfig}.
The first round starts with the server gathering information from all agents, followed by using the collection of agent information
to obtain an independent reply and send back to each agent in the second round.
The third round is a commit stage, where the devices upload encrypted and masked model renewals to the server.
Popular encryption techniques include public-key encryption~\cite{xu13pub}, homomorphic encryption~\cite{skeith07},
and Shamir's secret sharing schemes~\cite{wang17}.
The fourth round is the finalization stage where the devices release abundant ciphers to enable the server
to decrypt the collected model renewals.

In a DL system, local data can be exposed to an honest-but-curious server.
The authors of \cite{phong18} apply additively homomorphic encryption techniques to the global synchronization of the ML model to guarantee that the model is cryptographically safe against the server, since the usage of encryption can add tolerable overhead to a typical DL system.
Homomorphic encryption is an encryption technique which maintains a specific algebraic relation between the plaintext and ciphertext with a fixed encryption key~\cite{schoe11}.
A privacy-preserving DL system is presented, where many learning participants perform DL using a mixed dataset of all without the need for releasing their own data to the particular server.
The authors of \cite{phong18} address the problem of data leakage by constructing a strengthened system with the following attributes:
i) no data leaks to the server,
and ii) accuracy is maintained perfectly.
Asynchronous SGD is utilized to facilitate data parallelism and model parallelism.

\begin{figure}[t]
\includegraphics[width=0.43\textwidth]{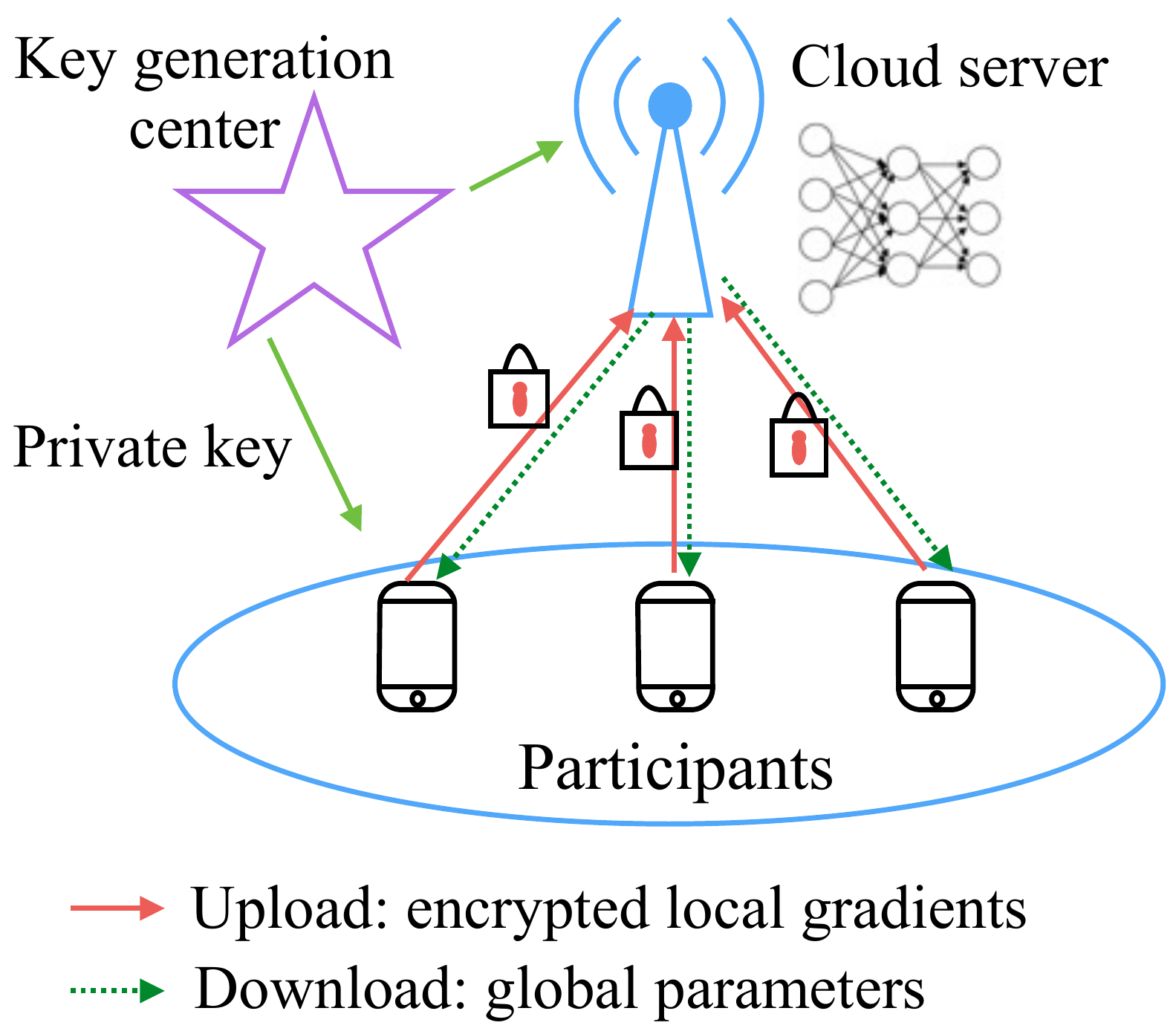}
\caption{An illustration of the system architecture \cite{hao20}.
The key generation center produces the public keys and releases the private keys to the cloud server and each participant,
the cloud server encrypts local gradients generated by the participants, and the participants train their local model over a private dataset.
}
\label{hao20syst}
\end{figure}

\subsection{Techniques and Implementation}

\subsubsection{Prevention of data poisoning}

Data poisoning is one of the cyberattacks very destructive to ML,
where an attacker injects ``poisoned'' samples into the datasets.
Any poisoned samples can be typical examples in regards to the mutual characteristic density in the domain, which is mislabeled.
They could also be examples that are atypical to the domain~\cite{miller20}.
For instance, if the sample is the transmit power of a BS, a user's location would be atypical.
Numerical results show that the distributed support vector machine (DSVM) is less prone to attacks
in a network with several nodes and a higher number of degrees.
The capability of DSVM against the attacks also depends on the network architecture and attack intensities.

Zhang and Zhu \cite{zhang20, Zhang2018A} propose secure decentralized techniques to safeguard  learning against data poisoning, as well as other system attacks.
A zero-sum game is designed to formulate the conflicting objectives between a legitimate learning agent who utilizes DSVMs and an attacker who can change sample data and labels.
The game characterizes the contention between the legitimate learner and the attacker.
A fully distributed and iterative algorithm is developed based on the ADMM technique to procure the instantaneous responses of the agent at every individual node to hostile activities.
In \cite{Zhang2018A}, the convergence of the decentralized approach is proved with no assumptions
on the sample data or network topologies.

\subsubsection{Perturbation plus encryption}
An FL algorithm is proposed in \cite{hao20} to protect user privacy and enhance system performance,
which could be used for industrial applications.
The system includes the key generation center that produces the public keys
and releases the private keys to the cloud server and each participant,
a cloud server which encrypts local gradients generated by the participants,
and the participants who train their local model over a private dataset,
as depicted in Fig. \ref{hao20syst}.
This privacy-preserving FL scheme is implemented by the following steps.

Each participating agent first disturbs the vector of its local gradients by applying a decentralized Gaussian technique.
This achieves the differential privacy.
Next, the disturbed gradient vector is encrypted into the Brakerski-Gentry-Vaikuntanathan (BGV) homomorphic encryption~\cite{BGV14}  ciphertext (called ``internal ciphertext''),
which is further nested into Augmented Learning with Error (A-LWE)~\cite{bansaka15} ciphertext
(namely, ``external ciphertext'') to realize secure gathering mechanism.
The A-LWE permits one to conceal auxiliary data into the false element.
Only after gathering ciphertexts from the agents can the server decode the external ciphertext correctly.
At the same time, the internal ciphertexts are added.
The server is able to decode collected values
without surrendering the privacy of individual agent.

Simulation results demonstrate that the privacy budget has little influence on the precision of the convolutional neural network (CNN) using this scheme.
However, the accuracy of the CNN drops dramatically when the collusion ratio is more than $0.5$,
which is the number of compromised participants over all of the participants.
The algorithm has been tested on the MNIST dataset, and its performance on high-dimensional datasets needs further qualification.

\begin{figure*}
\centering
\includegraphics[width=0.8\textwidth]{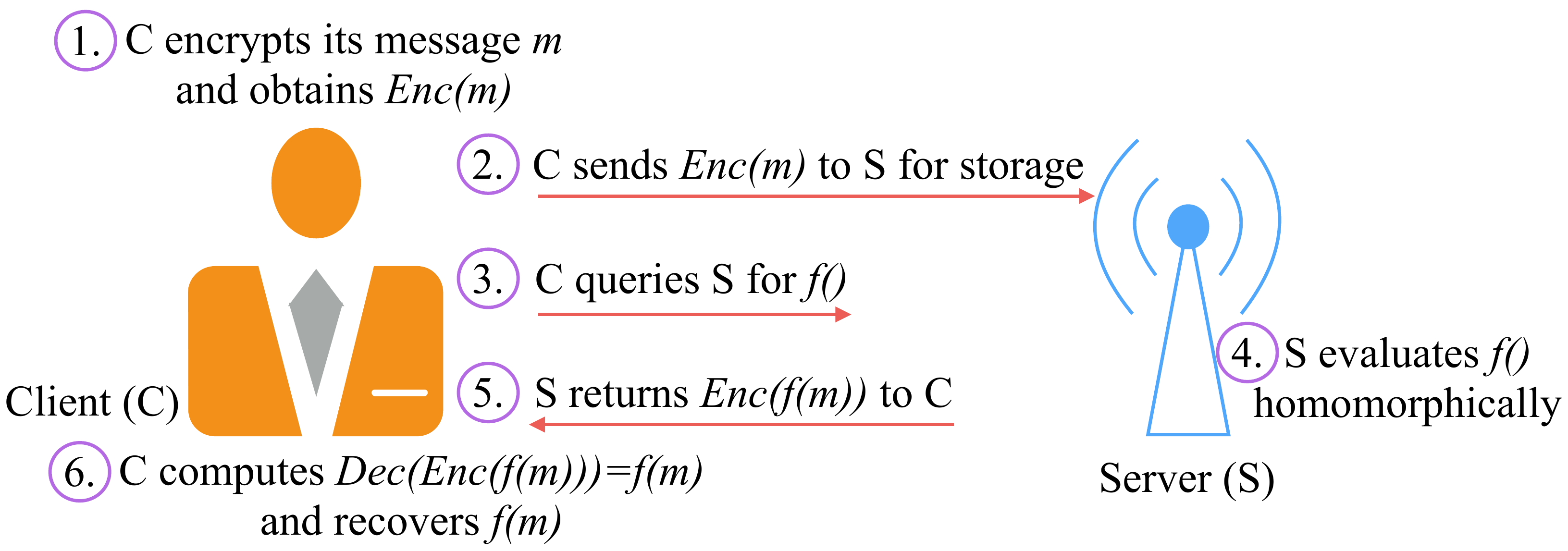}
\caption{An example of the homomorphic encryption in a client-server scenario,
where a client, C, initially encodes its private information (Step 1), then transmits the encoded information to the cloud server, S,
for storage (Step 2).
When C desires to operate (or calculate) a function (i.e., query), $f()$, over its individual information,
C transmits the function to S (Step 3).
S carries out a homomorphic manipulation over the encoded information, i.e.,
calculates $f()$ blindfolded (Step 4) and sends back the encoded outcome to C (Step 5).
Eventually, C retrieves the information via its individual secret key and procures $f(m)$ (Step 6)~\cite{acar18}.}
\label{homoenc}
\end{figure*}

\subsubsection{Blockchain-based solutions}
Blockchain is a recent distributed, anti-tampering ledger system initially designed in bitcoin and other cryptocurrency and later increasingly applied to the IoT~\cite{yusaber20, yugs20},
where a record of transactions is maintained distributively in a peer-to-peer network~\cite{yu2020, makh20}.
In \cite{kim20}, a blockchained FL-based framework is developed for the verification and transmission of a learning model
to achieve trusted decentralization between multiple devices.
An existing consensus mechanism of blockchain,
i.e., Proof-of-Stake (PoS)~\cite{yusaber20} or Byzantine-fault-tolerance (BFT), is deployed for the decentralization of training data.
The block generation algorithm can reduce the latency by adjusting the generation rate and Proof-of-Work (PoW) difficulty,
contributing to the minimization of the loss function.
A possible drawback of the algorithm is its susceptibility against a low SNR,
and the latency could increase dramatically with the decrease of the SNR.

Highly efficient BFT algorithms are designed to come up with correct and consistent decisions in a decentralized network in presence of some adversaries displaying malicious behaviors. This problem is known to be the Byzantine generals problem.
In \cite{Yang2018byrdie} and \cite{Yang2019byrdie}, a Byzantine-resilient algorithm for distributed learning, named BYRDIE, is developed to avoid sharing primitive data between workers, so that ML tasks can be accomplished in a completely distributed manner, even under the circumstances where Byzantine failures exist in the network.
With mild assumptions on the loss function (as the minimum of a convex combination of local empirical risk functions), BYRDIE splits the empirical risk minimization problem into multiple one-dimensional (scalar-valued) subproblems by using coordinate descent. Next, BYRDIE uses the Byzantine-resilient approach to solve each scalar-valued subproblem. The steps of BYRDIE involve an outer loop based on coordinate descent, and an inner cycle which achieves the solution to the scalar-valued optimization problem per cycle and strong tolerance to the Byzantine failures.

As discussed in \cite{Yang2019byrdie}, BYRDIE is an inexact distributed variant of coordinate descent, which can scale with the dimensionality of distributed learning problems. As a result, BYRDIE can address the challenge that the dimension of the training dataset is often substantially large, as compared to the neighborhood size of a node. (The minimum neighborhood of a node depends on the number of dimensions.)
By incorporating a local scrutinizing process into every iteration of the celebrated distributed coordinate descent technique by the network nodes, BYRDIE can detect Byzantine failures~\cite{Yang2018byrdie}.
In addition, BYRDIE improves the lowest of the statistical risk under the assumption that the typical training is based on i.i.d. datasets.



\section{Lessons Learned and Open Issues}

\subsection{Lessons Learned}
The main lessons learned can be summarized as follows.
\begin{itemize}
\item[$\bullet$] \emph{DML frameworks: }
Popular DML frameworks, such as FL and partitioned learning,
allow wireless devices to acquire a global model with little to no data exchange or based on partial models and datasets.
This can effectively protect data privacy and reduce communication cost~\cite{jakub16, CMFL19}.
Integrated with Q-learning or DL techniques, FL has a wide range of applications, for example,
power control~\cite{cao19, tao20}, QoS provisioning~\cite{haba19, yang20},
and spectrum management~\cite{kaibin19}.
In contrast, partitioned learning is the most welcome under edge computing settings~\cite{wen20}.
All the distributed learning techniques introduced in this paper allow for dataset partitioning, can reach the global optimality,
but may not converge fast.
Partitioned learning has model partitioning capability,
while federated RL does not need previously-stored dataset for model training.

\item[$\bullet$]  \emph{Parallel and distributed ML algorithms: }
Multiple variants of the SGD algorithm, such as asynchronous SGD, parallel mini-batch SGD and decentralized parallel SGD,
are described in~\cite{shokri15}.
These SGD-based techniques can train a large-scale model with fast convergence in a parallel and distributed fashion,
without sharing datasets among agents.
ADMM is an option in addition to SGD for DL or partitioned learning problems~\cite{anis20}.
Unlike SGD, ADMM can effectively prevent the disappearance of the gradient, and is resistant to poorly-qualified input data.
Global root-linear convergence can be achieved by the ADMM algorithm if the objective function is a combination of a smooth and strongly convex function and a non-smooth $\ell_1$ regularizer \cite{hong17}.

\item[$\bullet$] \emph{Privacy protection: }
Data poisoning and leakage are two serious threats to the security of ML systems,
especially in DML for wireless communications~\cite{ma20}.
To protect data privacy, techniques, such as perturbation, dummy, encoding,
and blockchain-based approaches, can be employed at the agent side,
while aggregation and secure multi-party computation (SMC) can be utilized at the servers.
Perturbation injects artificial noises to the clients' data.
Given a sufficient number of engaging agents, the perturbation can preserve client-side differential privacy at a little cost of model performance~\cite{geyer17}.
Dummy methods send dummy parameters, along with the genuine data, which bears redundancy. Therefore, they
require extra bandwidth and buffer, and consume more energy.
As a result, it is critical to strike a balance between augmented dummy communication cost and privacy degree~\cite{dummy16}.
Encoding encrypts the collective features of data and sends the features to the server, rather than the raw data~\cite{Zhang2019Privacy}.
Perturbation plus encryption is effective in preventing curious servers, especially when the compromised agents are fewer than
$50\%$ of all agents~\cite{hao20}.
Blockchain-based approaches utilize PoS or BFT to reach a consensus among the clients to achieve consistent DML~\cite{kim20}.
Data aggregation enables the server to extract and make full use of data in each update.
To guarantee system performance, it is important to select agents with high-quality parameters or easily-satisfied demands~\cite{gu19}.
SMC is a four-round encryption-based technique which prevents data inspection of the server~\cite{phong18}.
Since encryption adds overhead to the learning system, it is necessary to balance among the communication efficiency,
privacy level, and learning accuracy.

\item[$\bullet$] \emph{ML for Communication (MLC) vs. Communication for ML (CML): }
DML has been increasingly investigated to improve the performance of wireless communications, and it has become a branch of MCL \cite{2020Toward}. A key application is that DML deployed at edge networks of wireless or wired systems can optimize the physical and computing resource allocation to support emerging services, such as augmented reality (AR) and autonomous driving.
On the other hand, how to use DML at the network edge under communication and on-device resource constraints opens up a new research direction, i.e., CML \cite{2019Wireless}. In particular, edge ML architectures and their operations should be optimized under various on-device constraints (such as computing, memory and energy resources) and wireless communication limitations (such as communication overhead, channel dynamics and limited bandwidth).
CML and MLC are highly interrelated.

\item[$\bullet$] \emph{Architecture:}
Three typical DML architectures have been described in this paper. It is easy to use iterative MapReduce \cite{White12Hadoop}, as it originates from well-developed processing systems for big data. However, iterative MapReduce systems can only support synchronous communications. Therefore, the computing efficiency would be penalized, if implemented in a large machine cluster of heterogeneous machines with different computing capability, which is known as the straggler problem~\cite{amiri19dec}. On the other hand, the PS architecture supports both synchronous and asynchronous communications. It provides flexible interfaces to developers, through which programs on a single machine can be readily parallelized and run in a distributed fashion. However, the PS does not provide built-in ML algorithms. The graph-based architecture captures computations and communications on a graph. It also supports asynchronous communications. Moreover, it accommodates both data parallelism and model parallelism. One can choose from the existing architectures to fit the purpose, based on the specific ML problem (e.g., the model and data scales) and the characteristics of the machine clusters (e.g., the heterogeneity of the workers and the network bandwidth).

\end{itemize}
\subsection{Open Issues}
\begin{itemize}
\item[$\bullet$]  \emph{Meta-learning: }
Existing works focus on the training of traditional ML models and parameters with large-scale datasets.
However, these ML algorithms can be inefficient and even give incorrect answers when the dataset has fewer samples.
In this sense, it is critical to apply most recent ML techniques to training wireless networks,
such as zero-shot or few-shot learning~\cite{zeroshot, oneshot, fewshot}, and meta-learning~\cite{metaeffect}.
Zero-shot learning allows the system to distinguish a feature without specific training on it,
but rather training on data with relevant features.
Meta-learning empowers the system with the ability to learn and design an algorithm on its own,
which reduces human intervention.
The few-shot learning and meta-learning frameworks have been widely used in image recognition, text processing,
and robotic control,
and often been applied coupled with  DL, RL, and FL techniques~\cite{metasurvey, metaimage, fewtext, robot}.
It is an interesting and potentially rewarding research direction to extend zero-shot or few-shot learning and meta-learning frameworks
to wireless networks with limited CSI and data traffic samples.

\item[$\bullet$] \emph{Domain adaptation: }
Most of the existing works assume that the training and testing data have identical distributions of feature spaces,
and use the testing datasets to examine the learning effect.
Yet, in practice, the examination scenarios may vary and become uncontrollable.
The testing datasets may differ significantly from the training datasets in terms of feature space,
leading to underfitting and undesirable performances.
To address these issues, the technique called domain adaptation is applied
to map the non-i.i.d data from the source and target domains into a feature space based on the shortest distance criterion~\cite{domain}.
With the help of domain adaptation, the objective function trained on the source domain can be transferred to the target domain,
improving the learning accuracy in the target domain.
Domain adaptation can have the potential in distributed wireless networks, where the system only needs to collect (non-i.i.d) data from
some BSs while all the BSs can benefit from the training to improve their operation.
How to integrate this technique to effectively capture the massive time-varying data across devices
remains a challenge worth addressing in the future.

\item[$\bullet$] \emph{Anomaly detection: }
By training a model with a relatively large amount of consistent data samples and a very small number of abnormal samples,
a learning system is expected to distinguish anomaly (or deception) if some input data does not share the same feature as the training data.
Anomaly detection can be useful to monitor wireless networks for data traffic and CPU load,
and help detect anomalies in system operations.
Despite having been studied thoroughly in financial and industrial applications and systems~\cite{anomabrz, anomasg},
this issue is yet to be well investigated in the context of DML for wireless communications.
This is because the information in communication networks can vary rapidly over a short period,
resulting in a massive amount of data exchange.

\item[$\bullet$] \emph{Data provenance and model explainability: }
ML programmers prefer a simple, homogeneous, and consistent dataset as their input. When the analysis of large volumes of data is required, the entire dataset can hardly be handled by a single server and has to be placed in distributed file systems. Given the increasing demand for explainable models, not only do the programmers need to consider the ML algorithms, but analyze the distributed characteristics of the stored data and the effect of data pre-processing operations as well. To explain how a DML algorithm gives a decision, all transformations applied to the data should be considered.
It is claimed in \cite{Scherzinger2019The} that even basic transformations in data pre-processing, such as data partitioning, local data cleaning and value imputation, can have a strong impact on the resultant model. The effect becomes more apparent under a distributed setting. Tracing data provenance is a method to record transformations applied to the raw data. There are challenges and opportunities in linking the explainability of ML models and data provenance.

\item[$\bullet$] \emph{Architecture: }
This survey has introduced three mainstream architectures with their pros and cons. For example, the platforms based on the MapReduce architecture are more accessible, while their computing efficiency is unsatisfactory due to the straggler problem. What should the next-generation of DML architectures be? An architecture combined with AutoML is envisioned to be adaptive and make most of the pros of different architectures in the DML processes. This can be an interesting research direction to design future DML architectures.

\item[$\bullet$] \emph{Distributed DRL framework: }
For centralized DRL, the central controller equipped with abundant computational resources is able to perform training over a large volume of data for neural networks, at a high cost of spectrum usage and extended latency in wireless networks. Nevertheless, in a distributed DRL framework, end IoT devices usually have limited computational capacities and operate under constrained conditions or environments. This necessitates a holistic design of the distributed DRL framework, which is expected to decouple data perception, information sharing, and neural network training from RL algorithms at different devices~\cite{yuchao20}.
The overhead of information exchange among different entities of the distributed DRL network also needs to be taken into consideration.
Tremendous amount of effort has been devoted to decentralizing ML models, for example, FL.
On the other hand, multi-agent RL develops a game-theoretic interpretation of distributed operations and learns adequate strategies from the changes in the environment and the peers' actions.
It is poised to many distributed applications to wireless resource allocation, protocol coexistence, and cooperative communications.

\item[$\bullet$] \emph{Adaptation to network dynamics:}
Most of the existing distributed DRL frameworks are customized for individual networks, where the network states are relatively static in many cases. For future dynamic mobile networks with heterogenous agents, fast-changing network conditions and task requirements, challenges arise in how to ensure fast convergence and stable strategies. It is an open issue to design cooperative and coordinative policies between the agents for the best task execution performance, since complicated cooperations among agents can increase the state space and slow down the convergence.

\item[$\bullet$] \emph{Dataset generation: }
DML requires large datasets for both neural network training and performance analysis. Unlike traditional DL scenarios with referential data pools~\cite{maoq18}, in wireless systems, datasets are often generated artificially, e.g., by mathematic models. The synthetic datasets are simplifications of practical counterparts and may not be applicable in the real world. To this end, a new way of dataset generation for wireless networks is required to narrow down the gap between the emerging DML framework and practical systems.

\end{itemize}

\section{Conclusion}
We have provided a comprehensive survey of recent DML techniques applied to, and empowered by, wireless communication networks. Interesting applications of DML, including power control, spectrum management, user association, and edge cloud computing, have been discussed. The optimality, scalability, convergence rate, computation cost, and communication overhead of DML have been analyzed. We have also discussed the potential adversarial attacks that DML encounters, and the promising countermeasures. The lessons learned in this survey have been summarized. In an attempt to integrate DML into wireless systems design for flexibility and efficiency, interesting future directions have been outlined.


\bibliographystyle{IEEEtran}
\bibliography{distml_ref}

\balance

\end{document}